\documentclass[10pt, conference, compsocconf]{IEEEtran}

\usepackage{tikz}
\usepackage{mathptmx}
\usepackage[normalem]{ulem}
\usepackage{multicol}
\usepackage{pifont}
\usepackage{multirow}
\usepackage{color}
\usepackage{listings}
\usepackage{float}
\usepackage{ragged2e}
\usepackage{xspace}
\usepackage{soul}
\usepackage{lipsum}
\usepackage{makecell}
\usepackage{mathtools}
\usepackage{subcaption}
\usepackage[final]{microtype}
\usepackage[keeplastbox]{flushend}
\usepackage{listings}
\usepackage{multirow}
\usepackage{wrapfig}
\usepackage{etoolbox}
\usepackage{amsmath,amssymb,amsfonts}
\usepackage{algorithmic}
\usepackage{graphicx}
\usepackage{textcomp}
\usepackage{xcolor}

\usepackage[ruled,vlined,linesnumbered,algo2e]{algorithm2e}

\usepackage{fancyhdr}
\usepackage[hyphens]{url}
\usepackage[sort,nocompress]{cite}
\usepackage[bookmarks=true,breaklinks=true,letterpaper=true,colorlinks,citecolor=blue,linkcolor=blue,urlcolor=blue]{hyperref}

\newcommand{\proj}{\benchmark{ANT}\xspace}
\newcommand{\Fig}[1]{Fig.~\ref{#1}}

\newcommand{\Tbl}[1]{Tbl.~\ref{#1}}
\newcommand{\Sec}[1]{Sec.~\ref{#1}}

\newcommand{\bluec}[1]{{\textcolor{blue}{{#1}}}}

\newcommand{\benchmark}[1]{{\texttt{#1}}}
\renewcommand{\paragraph}[1]{\vspace*{0.15cm}\noindent\textbf{#1}\hspace*{.1cm}}

\newcommand{\ra}[1]{\renewcommand{\arraystretch}{#1}}

\newcommand{\good}{{\color{green}\checkmark}}
\newcommand{\bad}{{\color{red}\ding{55}}}

\makeatletter
\newcommand{\removelatexerror}{\let\@latex@error\@gobble}
\makeatother

\definecolor{myyellow}{RGB}{254, 230, 153}
\definecolor{myblue}{RGB}{193, 228, 248}
\definecolor{mygreen}{RGB}{199, 232, 172}

\newcommand{\revise}[1]{{\color{black}{#1}}}

\begin{document}
\IEEEoverridecommandlockouts
%
\title{ANT: Exploiting Adaptive Numerical Data Type for\\Low-bit Deep Neural Network Quantization
\thanks{$^1$ This work started during his internship at Microsoft Research.}
\thanks{$^2$ Jingwen Leng and Minyi Guo are corresponding authors of this paper.}
}

\author{
  \IEEEauthorblockN{
    Cong Guo\IEEEauthorrefmark{1}\IEEEauthorrefmark{2}$^1$,
    Chen Zhang\IEEEauthorrefmark{3},
    Jingwen Leng\IEEEauthorrefmark{1}\IEEEauthorrefmark{2}$^2$, 
    Zihan Liu\IEEEauthorrefmark{1}\IEEEauthorrefmark{2},
    Fan Yang\IEEEauthorrefmark{3},
    Yunxin Liu\IEEEauthorrefmark{4},
    Minyi Guo\IEEEauthorrefmark{1}\IEEEauthorrefmark{2}$^2$ and
    Yuhao Zhu\IEEEauthorrefmark{5}
  }
  \IEEEauthorblockA{
    \IEEEauthorrefmark{1}Shanghai Jiao Tong University, \IEEEauthorrefmark{2}Shanghai Qi Zhi Institute\\
    \{guocong, altair.liu\}@sjtu.edu.cn, \{leng-jw, guo-my\}@cs.sjtu.edu.cn
  }
    
  \IEEEauthorblockA{
    \IEEEauthorrefmark{3}Microsoft Research, \IEEEauthorrefmark{4}Institute for AI Industry Research (AIR), Tsinghua University \\
    chzhang1990@gmail.com, fanyang@microsoft.com, liuyunxin@air.tsinghua.edu.cn
  }
  \IEEEauthorblockA{
    \IEEEauthorrefmark{5}University of Rochester, yzhu@rochester.edu
  }
}
\maketitle

\begin{abstract} 
Quantization is a technique to reduce the computation and memory cost of DNN models, which are getting increasingly large.
Existing quantization solutions use fixed-point integer or floating-point types, which have limited benefits, as both require more bits to maintain the accuracy of original models.
On the other hand, variable-length quantization uses low-bit quantization for normal values and high-precision for a fraction of outlier values.
Even though this line of work brings algorithmic benefits, it also introduces significant hardware overheads due to variable-length encoding and decoding.

In this work, we propose a fixed-length \underline{a}daptive \underline{n}umerical data \underline{t}ype called \texttt{ANT} to achieve low-bit quantization with tiny hardware overheads. 
Our data type \texttt{ANT} leverages two key innovations to exploit the intra-tensor and inter-tensor adaptive opportunities in DNN models.
First, we propose a particular data type, \texttt{flint}, that combines the advantages of \texttt{float} and \texttt{int} for adapting to the importance of different values within a tensor.
Second, we propose an adaptive framework that selects the best type for each tensor according to its distribution characteristics.
We design a unified processing element architecture for \texttt{ANT} and show its ease of integration with existing DNN accelerators.
Our design results in 2.8$\times$ speedup and 2.5$\times$ energy efficiency improvement over the state-of-the-art quantization accelerators.     
\end{abstract}

\section{Introduction}\label{sec:introduction}
Deep neural networks (DNNs) have achieved great success in a variety of application domains, including computer vision~\cite{deng2009imagenet} and natural language processing~\cite{chowdhary2020natural}.
With the tensor-based computations as the dominant patterns, specialized tensor accelerators have been introduced for DNN inference~\cite{chakradhar2010dynamically, peemen2013memory,  chen2014dadiannao, du2015shidiannao, zhang2015optimizing, gokhale2014240, gupta2015deep } and training~\cite{jouppi2017datacenter,a100, qin2020sigma}.
However, the size of DNN models increases by $240\times$ every two years, significantly exceeding the hardware improvement rate ($3.1 \times$ every two years)~\cite{gholami2020ai}.
For instance, the recent large Transformer-based GPT-3~\cite{brown2020language} model has 175 billion parameters, whose single inference needs to take around 740 TOPs (tera operations).

Exploiting DNN model sparsity and redundancy through algorithm and hardware co-design is a promising way to overcome the widening computation gap between model and hardware.
There are generally two approaches, model pruning and model quantization, for reducing the computation and memory costs of DNN models.
First, pruning away the unimportant elements results in sparse DNN models with reduced parameter counts.
For example, NVIDIA has introduced the sparse tensor core since its Ampere architecture~\cite{a100}.
Second, model quantization uses narrow bit length to represent values to save memory and computation.
For example, Google's first generation TPU~\cite{jouppi2017datacenter} uses an 8-bit integer type for inference, while other commercial accelerators use the floating-point types with reduced precisions, such as FP16~\cite{a100}, TF32~\cite{a100}, and BF16~\cite{jouppi2017datacenter}, for accelerating the DNN training.

All the above algorithm and hardware co-design works for DNN models leverage the traditional numerical data types such as \texttt{int} and \texttt{float}, and these types are inherently inefficient for DNN models.
The reason is that the values in DNNs' tensors have both a non-uniform distribution and non-uniform importance.
For instance, many weight tensors in DNNs follow the Gaussian-like distribution, with many values around zero, which, according to many DNN pruning works~\cite{han2015learning, jung2019learning}, are not important and hence can be pruned.
However, the \texttt{float} type has the highest resolution (called rigid resolution~\cite{li2019additive}) for these small values, wasting its bit length.
On the other hand, the Gaussian-like distribution also has a long tail, whose range is critical for the DNN model accuracy~\cite{park2018energy, zadeh2020gobo}.
The \texttt{int} type needs a long bit length to represent the large values.
Given the above reasons, both \texttt{int} and \texttt{float} data types usually need more bits to maintain the original model accuracy~\cite{tambe2020algorithm, nagel2021white}.

To achieve even higher quantization benefits (i.e., lower bit length), prior works have proposed the outlier-aware quantization method and designed special hardware support~\cite{park2018energy, zadeh2020gobo}.
The basic idea is to employ the low precision 4-bit \texttt{int} for the small values with a high appearance frequency and high precision 32/16-bit \texttt{float} or \texttt{int} for large values with an extremely low frequency.
However, this method results in variable-length encoding and hence unaligned memory access, which are incompatible with the existing DNN accelerators and require a complex and high-cost hardware design.

In this work, we present an adaptive numeric data type called \proj{}, which can adapt to the importance of different value intervals within a tensor (i.e., intra-tensor adaptivity) and the distribution of different tensors (i.e., inter-tensor adaptivity).
More importantly, \proj{} has aligned memory accesses and efficient low-bit computation, leading to large quantization benefits and low hardware overheads.
We first propose a novel data type primitive called \texttt{flint} that combines the advantages of \texttt{float} with a large range and \texttt{int} with high precision for important value intervals.
We leverage the variable-length first-one coding technique to encode the exponent field.
As a result, the overall encoding has a fixed length, which is friendly for hardware decoding.

We then design a general framework that adapts to different distributions of weight and activation tensors, which include not only the above Gaussian distribution and also Laplace and uniform-like distributions.
We build upon the previous works like \texttt{PoT} (i.e., power of two) type~\cite{miyashita2016convolutional, zhou2017incremental}, and show how to integrate them into our \proj{} framework that chooses the best-fit numerical type to minimize the quantization error.
All those primitive types have the fixed-length, and a tensor can only have a fixed primitive type.
\revise{Compared to previous works that only exploit the intra-tensor adaptivity~\cite{sharma2018bit,park2018energy,zadeh2020gobo} or inter-tensor adaptivity~\cite{sharma2018bit,tambe2020algorithm}, the \proj{} framework can achieve both with high hardware efficiency.}

We further propose a unified TypeFusion processing element (PE) design that can handle the case when the input tensor and weight tensor have different primitive types.
The TypeFusion PE can be implemented on top of the original \texttt{float} or \texttt{int} multiply-accumulate (MAC) unit.
The required modification is a simple type decoder, which decodes different primitive types in \proj{} to a unified format for computing on the underlying \texttt{float} or \texttt{int} MAC unit.

The proposed \proj{} framework targets to solve the problem of the low-bit (i.e., 4-bit) quantization, which could still degrade the original model accuracy. 
We show that \proj{} is compatible with mixed-precision quantization~\cite{zhou2016dorefa, micikevicius2018mixed, wang2019haq, dong2019hawq, shen2020q, dong2020hawq, cai2020zeroq, cai2020rethinking, sharma2018bit, park2018energy, zadeh2020gobo, a100}.
In specific, our 4-bit \proj{} PE design can naturally support 8-bit \texttt{int} PE with little modification.
Owing to the mixed-precision support, \proj{} can use the 4-bit representation for over 90\% tensors and still maintain the same level of accuracy as the original full-precision models. 

We describe how to integrate the above \proj{} PEs into existing DNN accelerator architectures such as systolic array and tensor core~\cite{jouppi2017datacenter, a100}.
We present several optimizations to minimize the overhead of using \proj{}'s TypeFusion PE.
In particular, we show that \proj{} only imposes a simple type extension for the multiply-accumulate instruction, leaving the original programming model unmodified.
Our evaluation results show that the \proj{}-based accelerator surpasses the existing mixed-precision accelerator BitFusion~\cite{sharma2018bit} by 2.8$\times$ performance improvement and 2.5$\times$ energy reduction.

We make the following contributions in this paper.
\begin{itemize}
    \item We demonstrate the opportunities for adaptive quantization at both the intra-tensor level and the inter-tensor level, for which we present a unified qualitative framework to analyze the hardware overhead introduced by previous low-bit quantization works.
    \item We propose a composite adaptive numeric data type framework called \proj{} that can exploit the adaptive opportunities at both the intra- and inter-tensor levels in a hardware-friendly fashion.
    \item We propose a unified TypeFusion processing element (PE) design that can handle the case when the input tensor and weight tensor have different primitive types.
    \item We describe how to integrate the above \proj{} PEs to existing DNN accelerator architectures such as systolic array, which achieves the same level of accuracy as original full-precision models and significantly outperforms existing quantization accelerator under the same area.
\end{itemize}

\vspace*{0.3cm}
\section{Background}
\label{sec:background}
This section presents relevant background on DNN quantization, which has been widely studied to reduce the memory and computation cost of DNN models.

\revise{
\subsection{Quantization Metric}
\label{subsec:quant_metrics}
Many prior studies~\cite{zhang2018lq, choukroun2019low, banner2019post, nagel2020up} have shown that the optimization target of quantization is to reduce the MSE (Mean Square Error) between the original model and quantized model. 
In the digital image processing field~\cite{joshi2018digital}, the MSE metric is formally defined by the following equaiton:
\begin{align*}	
    MSE &=E[(x-{\hat{x}})^{2}] \\ & =\int(x-{\hat {x}})^{2}p(x)\,dx,
\end{align*}
where $x$ and $\hat{x}$ are the original and quantized value, respectively, and $p(x)$ is the the probability density function. 
 
To reduce the quantization MSE, it is natural to choose a numeric type whose quantization resolution distribution is similar to the tensor distribution~\cite{zhang2018lq, choukroun2019low, banner2019post, nagel2020up}.
Many researchers have proposed the distribution-aware quantization to address the non-uniform distribution, e.g., the Huffman encoding~\cite{han2015deep} and outlier-aware quantization~\cite{park2018energy, zadeh2020gobo}.
}

\subsection{{Fixed-length Quantization}}
\label{sec:tra}

Fixed-length quantization usually uses a narrow bit length representation based on \texttt{int} type or \texttt{float} type.
For example, there are works using 4-bit or 8-bit \texttt{int} types (called \texttt{int4} and \texttt{int8}, respectively) to quantize the weight tensors or activation tensors in DNN models. The \texttt{float}-based types can be represented by the following equation, with a varying exponent and mantissa filed. 
\begin{equation}
    \label{equ:float}
    \text{Real value} = \text{sign} \times 2^\text{exponent - bias} \times 1.\text{mantissa}
\end{equation}
For example, FP16~\cite{kahan1996ieee} uses a 5-bit exponent and 10-bit mantissa (5E10M), while BF16~\cite{jouppi2017datacenter} and TF32~\cite{a100} use the configuration of 8E7M and 8E10M, respectively.
There are also other more aggressive numerical types as follows.

\paragraph{PoT} type~\cite{miyashita2016convolutional, zhou2017incremental} (power of two) can be viewed as a special format of \texttt{float} type with only the exponent field and no mantissa field.
As a result, it can represent a large value range and its multiplication can be simplified to addition.

\paragraph{AdaptiveFloat} type~\cite{tambe2020algorithm} extends the basic \texttt{float} type to reduce the quantization errors for tensors with a non-uniform distribution (e.g., Gaussian-like distribution).
Its quantization framework adaptively sets the tensor-wise exponent bias to match the Gaussian-like distribution and reduce the MSE.

The quantization and dequantization function for a quantized element $\widehat{{w}}$ can be generalized to the following equation:
\begin{equation}
    \label{equ:quant}
    \widehat{{w}} = {s} \cdot {Dequant}[{Clamp} ({Quant}({\frac{{w}}{{s}}}), {min}, {max})]
\end{equation}
where ${s}$ is the quantization scale factor and, ${min}$ and ${max}$ are the lower and upper thresholds for the clipping function ${Clamp}(\cdot)$. The operator ${Quant}$ represents the quantization function. For example, the quantization function for \texttt{int} is a simple rounding-to-nearest function. After that, the dequantization operator ${Dequant}$ can decode the quantized number to the original type (e.g., 32-bit \texttt{float} or FP32), which is required for the quantization-aware training~\cite{jacob2018quantization}.

For memory-aligned quantization, we follow the common practice of per-channel weight quantization~\cite{nagel2021white}, which applies a separate scale factor for each output channel without additional hardware overhead.
For the input activations, we use the per-tensor scale factors because the per-channel activation quantization is challenging to implement~\cite{nagel2021white, li2021mqbench}.
These quantization granularities are widely used and supported by the DNN quantization frameworks such as TensorRT~\cite{TensorRT}. 
Moreover, we use the unsigned type to quantize activation tensors after ReLU efficiently~\cite{park2018energy,sharma2018bit,li2019additive,tambe2020algorithm}, as its outputs are all non-negative.
However, note that our framework supports both the signed numerical types and unsigned numerical types as we describe later.

\subsection{{Mixed-precision Quantization}}
Mixed-precision DNN quantization method uses different numbers of bits for a given data type to represent values in DNN tensors.
Many works~\cite{zhou2016dorefa, micikevicius2018mixed, wang2019haq, dong2019hawq, shen2020q, dong2020hawq, cai2020zeroq, cai2020rethinking, sharma2018bit, park2018energy, zadeh2020gobo, a100} have shown that the mixed-precision method is efficient for quantizing DNN layers that have different importance and sensitiveness for the bit length. 

The most widely used approach is tensor-wise mixed-precision, such as Bit Fusion~\cite{sharma2018bit} and NVIDIA's latest tensor core~\cite{a100}. 
In the tensor-wise quantization, all elements in each tensor use the same fixed-length numerical type. 
Thus, the tensor-wise mixed-precision is hardware-friendly without incurring much overhead, which has two kinds of implementation, i.e., temporal or spatial mixed-precision.
For example, a temporal design needs four cycles to perform an 8-bit $\times$ 8-bit multiplication using a 4-bit PE~\cite{song2020drq}, while a spatial design needs only one cycle using four 4-bit PEs~\cite{sharma2018bit}.

\subsection{{Outlier-aware Quantization}}
\label{sec:awareQ}
Outlier-aware quantization (OLAccel)~\cite{park2018energy} is tailored for Gaussian \revise{(non-uniform)} distribution, which is common in DNN models.
It divides the values in a tensor into two regions, i.e., outliers and non-outlier (or normal) values.  The outlier with a low probability can be represented by high precision (such as FP32 or FP16), and normal values with a high probability can be compressed with  fewer bits.
GOBO~\cite{zadeh2020gobo} is similar to OLAccel but has fewer outliers.
However, they exploit variable-length data encoding, which leads to the non-alignment in the memory sub-system.
As a result, these kinds of design increase the hardware complexity and have a non-negligible area overhead as we would show later.

\vspace*{0.2cm}
\section{Motivation: Adaptive Data Type}
\label{sec:motivation}
In this section, we first analyze the distributions of values in weight tensors and activation tensors from existing DNN models.
Prior works have proposed the accelerator microarchitecture with adaptive bit length to achieve low-bit quantization~\cite{park2018energy, zadeh2020gobo}, which requires a significant amount of hardware resources to deal with the variable length.
In contrast, our work proposes the idea of \underline{a}daptive \underline{n}umerical data \underline{t}ype (\proj{}) to fulfill the potential of low-bit quantization.
We present a qualitative framework to demonstrate the advantage of \proj{}, which is extremely hardware-friendly. 
 
\subsection{Opportunities for Adaptive Type}
\label{subsec:opportunity}
\begin{figure}[t]
    \begin{center}
    \includegraphics[width=1\columnwidth]{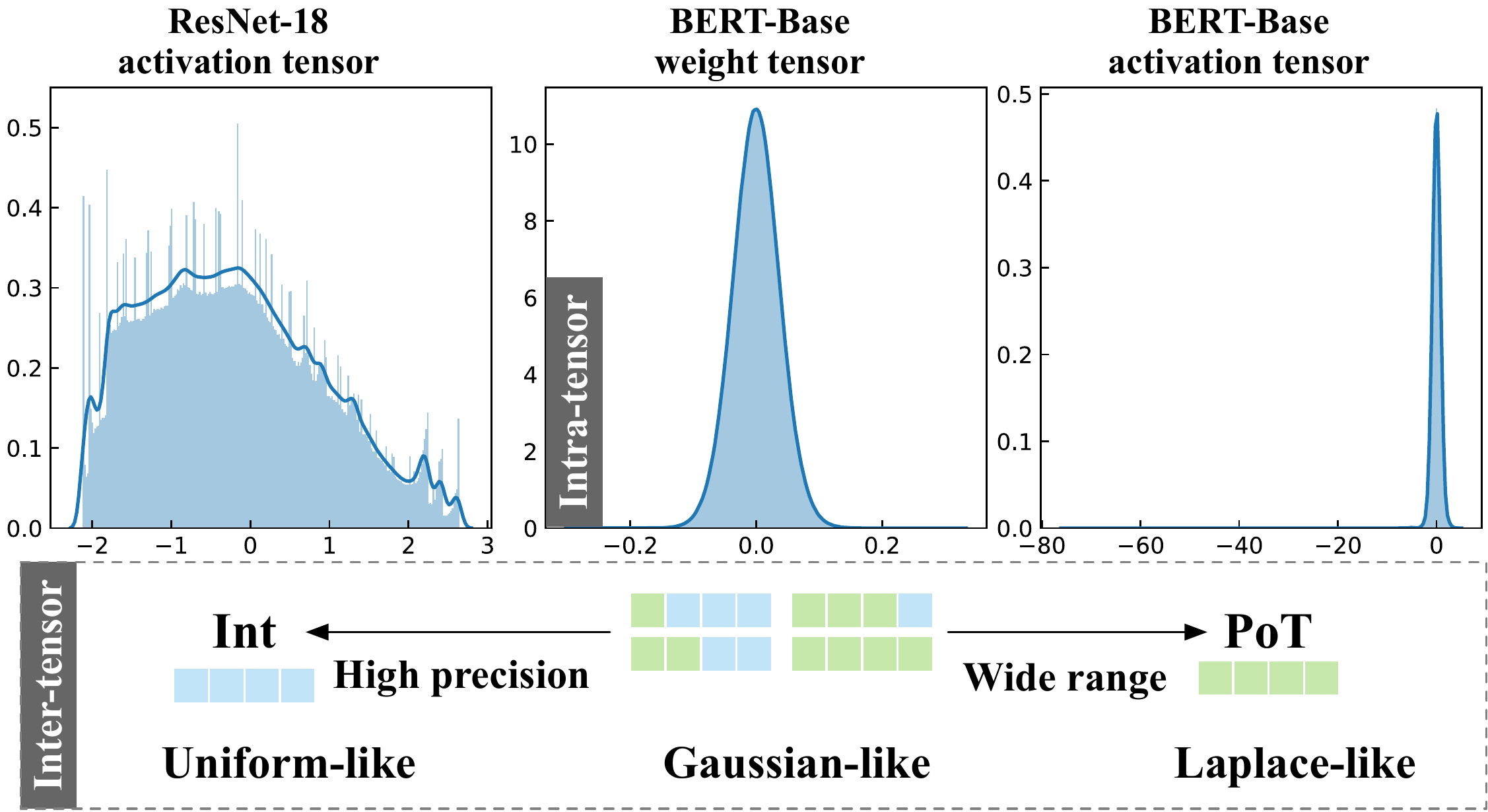}
    \caption{Intra-tensor and inter-tensor adaptivity.}
    \label{fig:tensor_d}
    \end{center}
\end{figure}

We identify two opportunities: \textbf{the inter-tensor adaptivity} to the specific distribution of each tensor and \textbf{the intra-tensor adaptivity} to the importance of values in each tensor.
These two opportunities lay the foundation for constructing the hardware-friendly adaptive quantization scheme. 

We first analyze the diversity of value distributions for various tensors in popular DNN models. 
Most previous works focus on Gaussian-based distribution and propose different techniques to mitigate the accuracy loss of quantization~\cite{park2018energy, li2019additive, zadeh2020gobo}.
However, tensors in DNN models exhibit different distributions as shown in \Fig{fig:tensor_d}.
For example, the activation tensor in the first layer of ResNet18~\cite{he2016deep} is closer to a uniform distribution, while an activation tensor in BERT~\cite{devlin2018bert} has a long tail that is close to Laplace distribution~\cite{banner2019post, bhandare2019efficient}.

\paragraph{Inter-tensor Adaptivity}
Given the diverse distribution of various tensors in DNN models, there naturally exists the opportunity called \emph{inter-tensor adaptivity}.
Intuitively, inspired by the non-uniform quantization of digital image processing~\cite{joshi2018digital}, if we adaptively choose the most suitable numerical type to quantize a tensor according to its distribution, it may achieve a lower quantization error, e.g., MSE (mean square error). 
For instance, in the left part of \Fig{fig:tensor_d}, the four-bit \texttt{int} type would lead to a smaller quantization error than the four-bit \texttt{float} type for the uniform-like distribution with a narrow range.
In contrast, the \texttt{PoT} type, which can represent a large dynamic range under the same bit length, is more suitable than \texttt{int} and \texttt{float} for the Laplace-like distribution with a long tail, as shown in the right part.

\paragraph{Intra-tensor Adaptivity}
Complementary to the inter-tensor adaptivity, the opportunity to reduce the quantization error also exists within a tensor. Specifically, extremely small and large values in a tensor do not require a high precision.
First, the key premise of DNN model pruning \revise{is that}~\cite{han2015learning, jung2019learning} small close-to-zero values are less important so can be pruned.
Second, many quantization works have shown that large values can be clipped to a threshold~\cite{choi2018pact}, which, however, should be large enough.
In other words, the exact numerical value of extremely large values is unimportant too as long as its rough numerical range is captured.
To exploit such an intra-tensor adaptivity opportunity, the quantization scheme should allocate fewer precisions for very small and large values while capturing a large enough range.

In the next part, we show that existing works cannot exploit the aforementioned opportunities, leading to marginal quantization benefits or significant hardware overheads.

\subsection{Quantization Architecture Analysis}
\label{subsec:analysis}

\begin{figure}[t]
    \begin{center}
    \includegraphics[width=1\columnwidth]{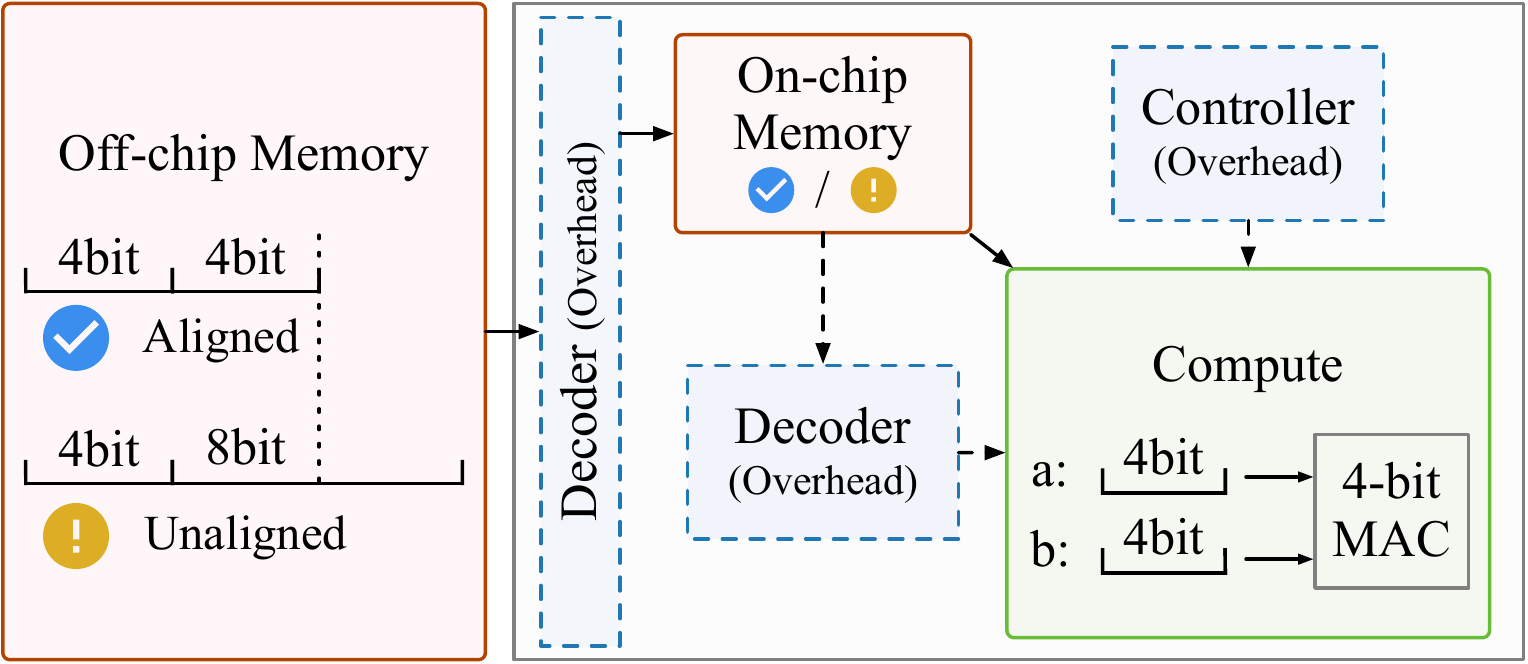}
    \caption{The qualitative framework for describing the hardware overhead for leveraging DNN quantization.}
    \label{fig:types}
    \end{center}
\end{figure}

We first present a unified qualitative framework in \Fig{fig:types} to analyze the hardware overhead introduced by previous quantization works.
We consider the three main components in the baseline DNN accelerator, which include the off-chip memory, on-chip memory, and computation unit.
For example, Google's TPU architecture~\cite{jouppi2017datacenter} has the off-chip HBM, large unified on-chip buffers, and weight-stationary-based systolic array as the computation unit. 
On top of the baseline architecture, a quantization scheme would generally introduce three extra components, which are off-chip data decoder, on-chip data decoder, and compute controller.

We analyze five state-of-the-art quantization schemes, which are \texttt{int}, AdaptiveFloat~\cite{tambe2020algorithm}, BitFusion~\cite{sharma2018bit}, \revise{BiScaled}~\cite{jain2019biscaled}, OLAccel~\cite{park2018energy}, and GOBO~\cite{zadeh2020gobo}.
\Tbl{tbl:quant_methods} compares their area overhead and quantization benefits with the metrics including averaged off-chip data width, averaged on-chip data width, and averaged compute data width.
For a fair comparison, we collect the statistics from over ten models including CNNs~\cite{simonyan2014very, he2016deep, szegedy2016rethinking}, ViT (vision transformer)~\cite{dosovitskiy2020image}, and BERT~\cite{devlin2018bert}.
We report numbers for all schemes when all models are close to their original \revise{FP32} accuracy (CNN with $<0.1\%$ loss and Transformer with $<1\%$ loss) \revise{except BiScaled}. 
\revise{We take the results from the BiScaled paper~\cite{jain2019biscaled}.
} 
\Sec{sec:evaluation} provides more experimental details.

\begin{table}[t]
    \ra{1.5}
    \resizebox{\columnwidth}{!}{%
    \begin{tabular}{l|c|c|c|c}
     \Xhline{1.2pt}
     \multirow{2}{*}{Architecture} & \multicolumn{2}{c|}{Off / On-chip Mem.} & \multicolumn{1}{c|}{Compute} & Overhead\\ \cline{2-4}
     & Aligned & Bit Width & Bit Width & Area Ratio \\
     \Xhline{1.2pt}
    Int & \good & 8 & 8 & $0$ \\ \hline
    AdaFloat~\cite{tambe2020algorithm} & \good & 8& 8 & $14.5\%$ \\ \hline
    BitFusion~\cite{sharma2018bit} & \good & 7.07& 7.07  & $\sim 0$ \\ \hline
    \revise{BiScaled}~\cite{jain2019biscaled} & \good & \revise{6.16}& \revise{6.16}  & \revise{7.1\%$^{**}$} \\ \hline
    OLAceel~\cite{park2018energy} & \bad & 5.81 & 4.36 & $71\%$ \\ \hline
    GOBO$^{*}$~\cite{zadeh2020gobo} & \bad & 4.04 / 6.81 &16 &  $55\%$ \\  \hline
    \textbf{ANT (Ours)} & \good & {4.23} & 4.23 & $0.2\%$ \\
    \Xhline{1.2pt}
    \end{tabular}%
    }
    \caption{ Quantization architecture comparison. We collect the average bit of once memory access and computation precision among 13 workloads, including CNN and Transformer, for each quantization method. We also count the area ratio of the decoder and controller. $^{*}$GOBO only has the weight quantization, where its statistics only involve weight tensor. \revise{$^{**}$We only synthesize the BPE area of BiScaled.}}
    \label{tbl:quant_methods}
\end{table}

The conventional \texttt{int}-based quantization stores tensors with the same bit width in both off-chip and on-chip memory, making them all access aligned.
Thus, it does not require any additional decoder logics and compute controller (i.e., its area overhead is zero).
On the other hand, it does not exploit the inter- and intra-tensor adaptivity.
A low-bit \texttt{int} can only represent a narrow range, which may clip away few but important large values~\cite{choi2018pact}.
As such, it often requires 8-bit \texttt{int} type to retain original model's accuracy.
Hence its quantization benefits are limited to 8 bit for off-chip memory, on-chip memory, and computation resources.

AdaptiveFloat~\cite{tambe2020algorithm} (shorted as \texttt{AdaFloat}) extends the \texttt{float} type with a tensor-wise exponent bias.
It has aligned off-chip and on-chip memory accesses but requires an exponent bias decoder for controlling the bias offset, whose area is 14.5\% larger than the fixed-point (\texttt{int}).
Although floating-point allows \texttt{AdaFloat} to represent a greater value range, it gradually increases quantization resolution as values' magnitude decreases logarithmically, leading to excessively high resolution (called rigid resolution~\cite{li2019additive}) for much smaller values. 
Because smaller values are usually less important, the rigid resolution wastes much numerical representation space, rendering its overall quantization benefits to 8 bits.

BitFusion~\cite{sharma2018bit} exploits the inter-tensor adaptivity by choosing different bit lengths (or precisions) for different tensors. 
It incurs an almost zero hardware overhead because the high precision data type (e.g., 8-bit \texttt{int}) can reuse all the low-precision data type (e.g., 4-bit \texttt{int}) components without extra overhead.
However, its underlying primitive data type is still \texttt{int} type, which limits its quantization benefits to 7.07 memory bits and computation bits on average.

OLAccel~\cite{park2018energy} and GOBO~\cite{zadeh2020gobo} leverage the outlier-aware quantization scheme.  
They store a tensor using a variable-length compressed form (e.g., 4 bits for normal values and 16 bits for outliers with relatively large values) in the off-chip memory, which requires a dedicated  data decoder.
They also require an additional outlier controller to orchestrate the computation between normal values and outliers. 
Note that GOBO~\cite{zadeh2020gobo} only supports weight quantization, it requires high precision (i.e., 16-bit) floating-point computations for activation tensors.
Even though both designs have a large quantization benefit with low memory bits, their associated hardware complexity and overhead are also significant.

BiScaled~\cite{jain2019biscaled} also adopts outlier-aware quantization but with a fixed-length compressed form. 
However, it requires an extra bit mask for indicating different scale factors, which leads to a more considerable area overhead. 
Moreover, it only considers two value ranges for the intra-tensor adaptivity, leading to the benefits of 6.16 memory and computation bits.

To balance the hardware overhead and quantization benefits, we propose the adaptive numerical data type (\proj{}) framework that exploits both inter-tensor and intra-tensor adaptivity. 
\proj{} further supports mixed-precision. 
As the last row in \Tbl{tbl:quant_methods} shows, \proj{} achieves the lowest average bit for memory and computation for both activation and weight tensors with a negligible area overhead.
\vspace*{0.2cm}
\section{Adaptive Numeric Data Type}
\label{sec:Overview}

In this section, we present our adaptive numeric data type \proj{} that can exploit both the intra- and inter-tensor adaptive opportunities in a hardware-friendly fashion.
We first present a novel primitive data type called \texttt{flint} that combines the advantages of \texttt{float} and \texttt{int} for adapting to the importance of different values within a tensor.
We then propose a general framework that adapts to each tensor's distribution by selecting different primitive types, including \texttt{int}, \texttt{float}, \texttt{flint}, and \texttt{PoT}.
As a result, \proj{} has aligned memory accesses and efficient low-bit computation, translating to significant benefits with low hardware overheads.

\subsection{Intra-tensor ANT: {Flint}}
\label{subsection:flint}

\begin{figure}[t]
    \begin{center}
    \includegraphics[width=1\columnwidth]{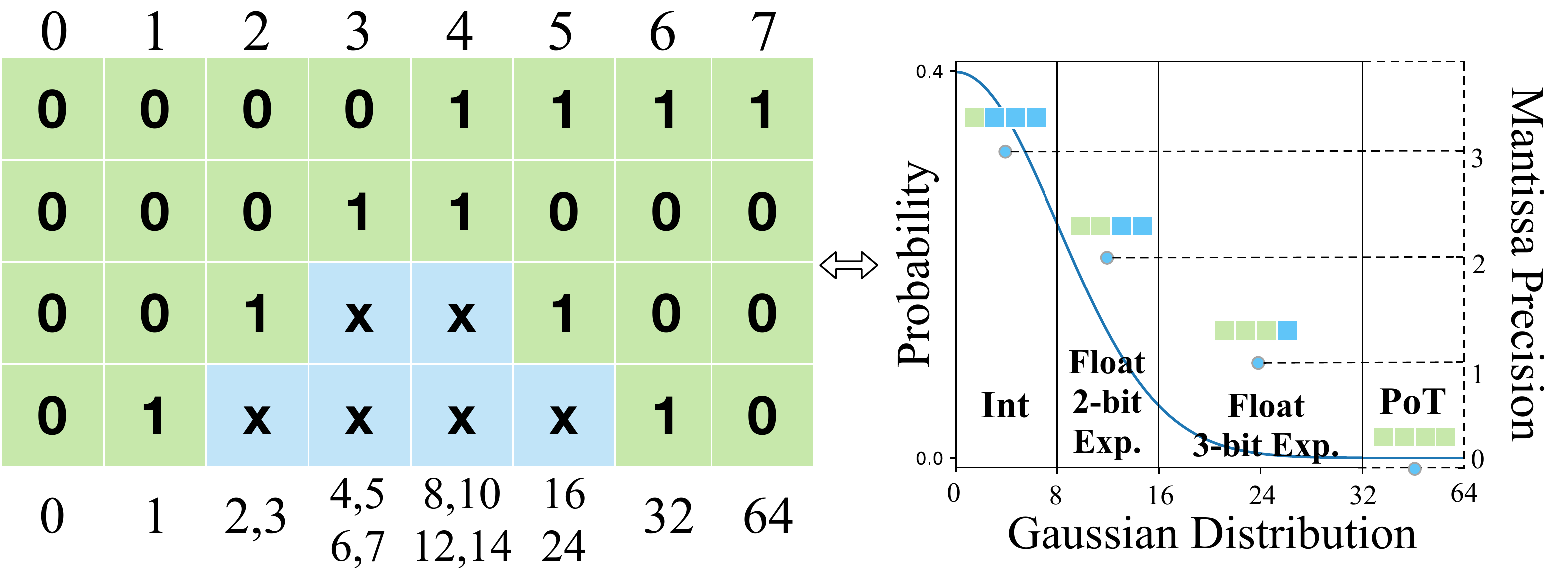}
    \caption{The 4-bit unsigned \texttt{flint} type, with ``x'' as either 0 or 1. The exponent and mantissa bits are marked with green and blue color, respectively.}
    \label{fig:flint}
    \end{center}
\end{figure}
\paragraph{Main Idea}
As shown in \Sec{subsec:opportunity}, extremely small and large values in a tensor do not need high precision. 
A naive approach is to divide a value range into multiple intervals and assign fewer bits to intervals with small or large values.
However, this leads to variable-length for different elements in the tensor and requires expensive hardware logic to handle the incurred unaligned accesses as mentioned in \Sec{subsec:analysis}.

To provide a fixed-length data type overcome while exploiting intra-tensor adaptivity,
we propose a new primitive data type called \texttt{flint}. 
Our main idea is to start with a fixed-length, and allocate fewer mantissa bits (i.e., more exponent bits) to extremely small and large values (as they do not require a high precision) while allocating more mantissa bits (i.e., fewer exponent bits) to middle-range values to preserve their precision. Using more exponents bits for large values also allows us to capture the range of very large values.

To mark the boundary between the exponent and mantissa field, we use the first appearance of bit `1' after the most significant bit (or sign bit). We call this encoding \textit{first-one encoding}.
While other strategies exist to split the exponent and mantissa fields, this encoding has the critical advantage of simplicity:
the decoder for this encoding only requires a simple leading zero detector as we would show later.

\paragraph{An Example} We use the example of four-bit \texttt{flint} in \Fig{fig:flint} to illustrate our design.
Without loss of generality, we assume the case of unsigned values that have been scaled with the per-channel (or per-tensor) granularity for the weight (activation) tensor as described in \Sec{sec:tra}.

 \Fig{fig:flint} left shows a four-bit unsigned binary number using our \texttt{flint} encoding, which can represent 16 distinctive binary values with the maximum value of 64.
We divide this value range to eight intervals corresponding into the eight columns in \Fig{fig:flint} left, and highlight the exponent fields in green color.
The first four intervals have the encoded exponent fields of \texttt{0000$_2$}, \texttt{0001$_2$}, \texttt{001$_2$}, and \texttt{01$_2$}.
Under the four-bit fixed-length encoding, the number of mantissa bits for these intervals are \texttt{0}, \texttt{0}, \texttt{1}, and \texttt{2}, respectively.
This mantissa bit allocation scheme is adaptive to the value importance as the first two intervals are closer to zero and hence have the least number of precisions.
The exponent-mantissa bit allocation for the last four intervals is inverse to the first four intervals.
In specific, the greatest interval \texttt{1000$_2$} has no mantissa bit, which is also desirable because the range is more important than the precision for large values.

\begin{table}[t]
    \centering
    \ra{1.3}
    \resizebox{\columnwidth}{!}{%
      \begin{tabular}{c|c|c|c}
        \Xhline{1.5pt}
        Bits   & Exponent Value & Fraction Value & Value in Decimal \\ \Xhline{1pt}
        \bluec{0000}& - &0&0 \\ \hline
        \bluec{0001}& $1-1 =0$&1&$2^0 \times 1 = 1$  \\ \hline
        \bluec{001}x& $2-1 =1$&1, 1.5& 2, 3  \\ \hline
        \bluec{01}xx& $3-1=2$&1, 1.25, 1.5, 1.75&4, 5, 6, 7\\ \hline
        \bluec{11}xx& $4-1=3$&1, 1.25, 1.5, 1.75&8, 10, 12, 14\\ \hline
        \bluec{101}x& $5-1=4$&1, 1.5&16, 24\\ \hline
        \bluec{1001}& $6-1=5$&1&32  \\ \hline
        \bluec{1000}& $7-1=6$&1&$2^6 \times 1 = 64$  \\ \Xhline{1.5pt}
        \end{tabular}
    }
        \caption{The value table of 4-bit unsigned \texttt{flint} with the exponent bias of $-1$. The blue numbers are the first-one-encoded exponent and ``x'' is mantissa with value of 0 or 1.}
        \label{tbl:ant_value}    
\end{table}

\Tbl{tbl:ant_value} shows the value table for the above 4-bit unsigned \texttt{flint} with the exponent bias of $-1$.
Each row refers to the divided interval (i.e., column) in \Fig{fig:flint} left.
The equivalent exponent value for a given \texttt{flint} encoding is the interval number plus the bias.
The final decimal value equals the fraction value multiplied by the exponent value raised by the power of two as in Equation~\ref{equ:float}.
For example, the \texttt{flint} encoded number \texttt{1110$_2$} has the exponent value of $4-1=3$ and mantissa bit \revise{\texttt{10$_2$}}, which corresponds to the fraction value of $1.5$.
As such, its decimal value is $2^{3} \times 1.5 = 12_{10}$.

Essentially, our proposed \texttt{flint} is a mixture of \texttt{int}, \texttt{float} (and its variants), and \texttt{PoT} at different intervals.
The first four intervals (i.e., rows) in \Tbl{tbl:ant_value} have the binary encoding of \texttt{0000$_2$}, \texttt{0001$_2$}, $...$, \texttt{0111$_2$} and represent the integer value of $0, 1, \dots, 7$, respectively.
In other words, \emph{the four-bit \texttt{flint} type is equivalent to \texttt{int} in the first four intervals.}
The 5th and 6th intervals have the 2 and 1 mantissa bits, making them equivalent to the \texttt{float} with 2 and 1 mantissa bits (i.e., 2 and 3 exponent bits), respectively.
The last two intervals have zero mantissa bits, which are equivalent to the \texttt{PoT} type.

Given these above insights, we call the proposed type \textbf{flint}, which is able to combine the advantages of \textbf{fl}\texttt{oat} and \textbf{int}.
Note that in the right of \Fig{fig:flint}, the divided eight intervals can be coalesced into four intervals according to their type-equivalence.
We can see that the mantissa precision allocation in \texttt{flint} highly matches the Gaussian distribution, meaning values with a higher frequency also with more mantissa bits.
As most tensors in DNN models are Gaussian-like, we show later that this behavior leads to fewer quantization errors.

\paragraph{Flint Encoding Algorithm}
To recover the accuracy loss, it is generally required to perform the fine-tuning with the quantization in the training loop.
As such, \texttt{flint} needs an encoding algorithm to convert the original high precision values, such as FP32, to the low precision \texttt{flint}. 
The software can use this encoding algorithm to mimic the \texttt{flint} behavior during fine-tuning.
Meanwhile, as we target both weight and activation quantization, the encoding needs to be performed dynamically during inference, which requires a lightweight and hardware-efficient encoding algorithm.

\begin{figure}[t]
    \begingroup
    \removelatexerror
    \begin{algorithm2e}[H]
        \DontPrintSemicolon
        \KwIn{
            Element, $e$;
            Bit-width, ${b}$;
            Scale factor, $s$.
        }
        \KwOut{
            Quantized Element, $q$.
        }
    
        \SetKwFunction{FMain}{FlintQuant}
        \SetKwProg{Fn}{def}{:}{}
        \Fn{\FMain{$e$, $b$, $s$}}{
            $en = 2\times b$;  \tcp*[h]{Exponent number.}\\ 
            $e = $ IntQuantization($e$, $s$, 0, $2^{en- 2}$); \\
            \If(){$e == 0$}{
                \Return{$0$};}
            \Else(){
                $i = \lfloor log_2(e) \rfloor + 1$; \\
                $exp = $ GetExponent($b, i$);  \tcp*[h]{First-one exponent.} \\
                $mb = b-$ len($exp$); \tcp*[h]{Mantissa bit.} \\
                $m = Round[(e / 2^{i-1} - 1) \times 2^{mb}]$; \\
                $m = Binary(m)$; \\
                $q = Concat(exp, m)$;\\
                \Return{$q$};
            }
        }
        \caption{Element-wise \texttt{flint} encoding algorithm.}
        \label{alg:flint_encoder}
    \end{algorithm2e}
    \endgroup
\end{figure}

Algo.~\ref{alg:flint_encoder} details the hardware-efficient \texttt{flint} encoding algorithm for each tensor element.
First, a $b$-bit \texttt{flint} number has $2\times b$ possible first-one codes for exponents (Line 2), so its value interval is $[0, 2^{2b-2}]$.
We first use \texttt{int} quantization with the scale factor $s$ to quantize the input value $e$ to its integer value with the value range $[0, 2^{2b-2}]$ (Line 3).
We then calculate its value interval index according to the interval boundary in \Tbl{tbl:ant_value} (Line 7), and derive the exponent and mantissa filed correspondingly (Line 8 - 12).

For example, the 4-bit unsigned \texttt{flint} type has the value range of $[0, 2^{2\times 4 - 2} = 64]$ (Line 2).
For a decimal number \texttt{11}$_{10}$, it has been quantized by the \texttt{int} quantization with the scale factor $s$ (Line 3).
We then calculate its value interval index $i = 4$ (Line 7), for which the encoded exponent is $11_2$ (Line 8).
After deriving the exponent field, we know its mantissa bit-width is $mb = 4-2=2$ (Line 9), with value of $m = (11 / 8 -1) \times 2^2 = 1.5_{10}$ and rounded to $m = 2_{10}$ (Line 10). 
This binary code of $2_{10}$ is $10_2$ (Line 11), which is concatenated with exponent $exp$ to get the final \texttt{flint} encoded number $q = 1110_2$ (Line 12).
Note that after the above quantization process, the original value $11_{10}$ is now rounded to $12_{10}$ in \texttt{flint} representation. 

The above \texttt{flint} encoding algorithm is an element-wise function that can be implemented efficiently in both hardware and software.
The exponent and mantissa bit settings are constants when the quantization bit-width is given. 
The quantization of weight tensors can be done offline, while  activation quantization needs hardware support.
Owing to the simplicity of our encoding algorithm, we can implement it in the hardware by augmenting the hardware's element-wise computation unit, such as the activation unit.

\subsection{Inter-tensor ANT}
To exploit the inter-tensor adaptivity and balance the hardware complexity, we propose to select the data type for a \emph{tensor} according to its distribution.
As we have shown previously, \texttt{flint} is equivalent to \texttt{int}, \texttt{float}, and \texttt{PoT} in certain value intervals.
\proj{} is natural to support these four primitive data types for inter-tensor adaptivity.

As previously shown in the right of \Fig{fig:tensor_d} and \Fig{fig:flint}, the \texttt{int} is most suitable for the uniform-like distribution.
The \texttt{float} or \texttt{PoT} are most suitable for Laplace-like distributions.
The \texttt{flint} data type is most suitable for Gaussian distribution because \texttt{flint} has the highest resolution (most mantissa bits) for the values with the highest frequency.
In our work, we propose an automatic algorithm to determine the data type for tensors in a trained DNN model that we describe later.
Meanwhile, it is hardware-efficient to support the above four data types.
Even with different data types, a tensor is stored in a fix-length format. 
As such, the memory accesses of \proj{} are aligned and hence efficient.

\begin{figure}[t]
    \begingroup
    \removelatexerror
    \begin{algorithm2e}[H]
        \DontPrintSemicolon
        \KwIn{
            Tensor, $T$;
            Candidate list of numeric types, ${L}$.
        }
        \KwOut{
            Quantization function, $F_Q$.
        }
    
        \SetKwFunction{FMain}{\proj{}}
        \SetKwProg{Fn}{def}{:}{}
        \Fn{\FMain{$T$, $L$}}{
            $minMSE = 10^9;$ \\
            \ForEach{$l \in L$} 
            {        
                $F =$ GetQuantFunc($l$); \tcp*[h]{Get the quantization method of $l$.} \\
                $m =$ ArgminMSE($T$, $F$); \tcp*[h]{Search the minimum MSE with range clipping.} \\
                \If{$m < minMSE$}
                {
                    $F_Q = F$;
                }
            }
            \Return{$F_Q$}
        }
        \caption{\proj{} data type selection algorithm.}
        \label{alg:ant}
    \end{algorithm2e}

    \endgroup
\end{figure}

\subsection{ANT-based Quantization Framework}
\label{subsec:ant_quantization}

To apply \proj{} for quantizing DNN models, we first need to select the specific type for a given tensor since \proj{} contains multiple primitive data types.
After that, we then perform the fine-tuning to recover the accuracy loss.
We describe the details in each step and explain how to use the \proj{}-quantized model for inference in the end.
    
\paragraph{Type Selection}
Algo.~\ref{alg:ant} shows the type selection algorithm for \proj{}, which chooses the primitive data type with minimum mean squared error (MSE) out of the candidate list $L$ (e.g., \texttt{flint}/\texttt{int}/\texttt{float}/\texttt{PoT}).
We first get the quantization function for each candidate type (Line 3-4).
The \texttt{flint} encoding algorithm is described previously in \Sec{subsection:flint}, and the de-quantization algorithm can be derived by inverting the process.
We use the original quantization function for the other primitive types.
For a given data type, we also need to determine its range (i.e., the scaling factor).
We employ a widely-used range clipping method~\cite{choukroun2019low, cai2020zeroq} that determines the clipping range by minimizing the MSE (Line 5). 
We then determine the most suitable data type for each tensor with minimum MSE from the candidate list (Line 6-7).

We only execute the above type selection algorithm once per tensor before fine-tuning.
The reason is that the distribution of tensors of a well-trained model remains roughly similar even during the fine-tuning stage~\cite{banner2019post}, which has been exploited by many other quantization methods~\cite{park2018energy,li2019additive,sharma2018bit}.
For the weight tensor quantization, we do not require any training samples and directly use the weight tensors from the original, trained DNN models to determine each weight tensor's data type.
For the activation tensor quantization, we need about 100 training samples to collect the statistical information for determining the types.

\paragraph{Mixed Precision} \proj{} is also compatible with the mixed-precision quantization method to achieve the same level of accuracy as the original DNN model. We leverage a layer-wise precision selection method~\cite{sharma2018bit}.
In the beginning, we use the 4-bit \proj{} type for all layers and perform fine-tuning.
We then collect and sort the MSE of all layers in descending order. 
We enlarge the bit width of a layer with the greatest MSE to 8 bits and then perform another fine-tuning.
We repeat the above process until the accuracy of the quantized model is within the preset threshold of the original model.

\begin{figure}[t]
    \begin{center}
    \includegraphics[width=1\columnwidth]{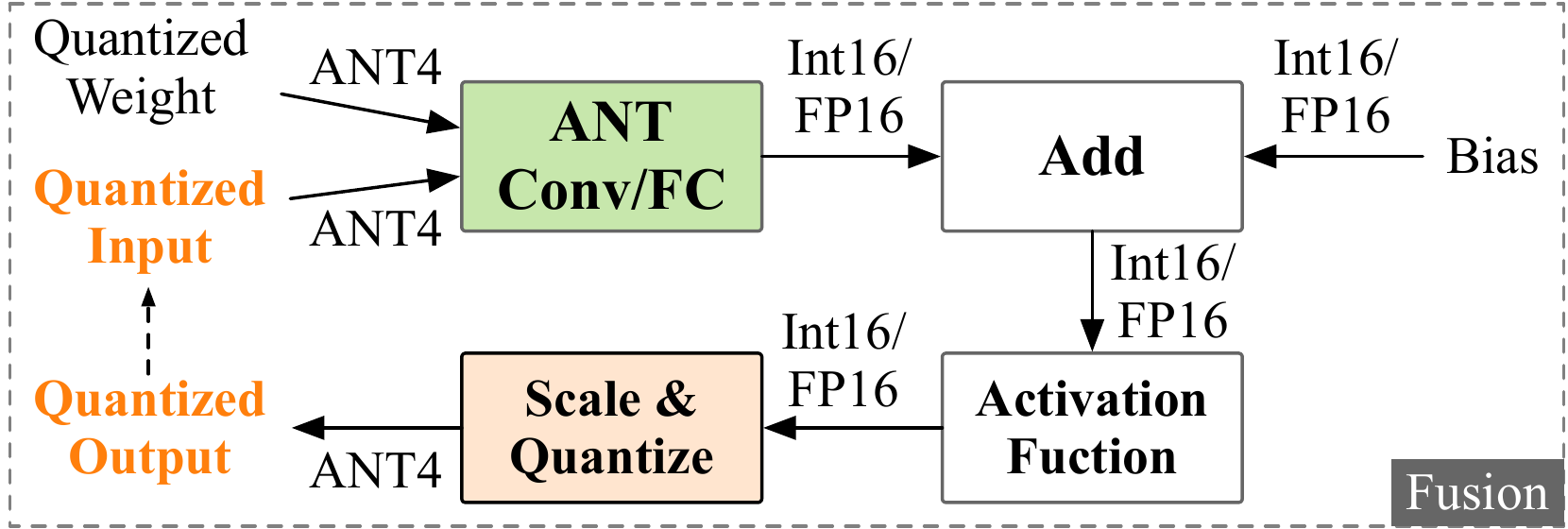}
    \caption{\revise{\proj{}-based quantized inference.}}
    \label{fig:ant_framework}
    \end{center}
\end{figure}

\paragraph{ANT-based Inference} 
\Fig{fig:ant_framework} presents the \proj{}-based inference framework.
For convolution and fully-connected layers, we use the low-bit quantized weights and input activations but keep the output activations the high precision.
The reason is two-fold.
First, the accumulation in these layers needs to maintain a high precision~\cite{jacob2018quantization}.
Second, their following layers are usually activation layers such as \texttt{SoftMax} and \texttt{GeLU}, which also require high-precision numbers~\cite{zafrir2019q8bert}.
The output tensors of activation layers can be quantized to low-bit values, which can be completed in the hardware by augmenting the activation units (or their equivalence).

\vspace*{0.2cm}
\section{Type-Fusion Processing Element}
\label{sec:typefusion}

The \proj{} data type introduces unique challenges for the design of the processing element because the PE now needs to handle different primitive types (\texttt{flint}/\texttt{int}/ \texttt{float}/\texttt{PoT}).
Moreover, the input activation tensor and weight tensor for the same layer may have different data types.
To address these challenges, we propose the TypeFusion processing element architecture that supports the multiply-accumulate (MAC) operation between different primitive types.
We describe the two cases where we build TypeFusion PE on top of the original \texttt{float}-based PE and \texttt{int}-based PE, respectively.
For convenience, we simplify the description below with the focus on unsigned numbers and it is straightforward to adapt the described design to support signed numbers.

\subsection{Float-based PE}

We first describe how to augment the original \texttt{float}-based MAC unit to support \texttt{int},  \texttt{PoT}, and \texttt{flint}.
As we have described previously in \Sec{subsec:ant_quantization}, the accumulation needs to be performed in high precision, which is sufficient to cover the ranges of low-precision \proj{}. 
As such, we focus on how to augment the multiplication component.

\paragraph{Multiplier} 
For the \texttt{int} and \texttt{PoT}, we can regard them as two special \texttt{float} formats. \texttt{Int} has no exponent and is full of mantissa with the subnormal number. 
\texttt{PoT} has no mantissa and is full of exponent bits with extreme dynamic range. For each type, we need to identify its exponent bit-length and mantissa bit-length and send them to exponent and mantissa decoders, respectively. Therefore, we need a \texttt{float} multiplier with an $n$-bit exponent and an $n$-bit mantissa for $n$-bit \texttt{int} and \texttt{PoT}.
Meanwhile, those exponent and mantissa bits are sufficient for the $n$-bit \texttt{flint}, which is equivalent to \texttt{float}, \texttt{int}, and \texttt{PoT} in different value intervals.

\begin{figure}[t]
    \begin{center}
    \includegraphics[width=0.9\columnwidth]{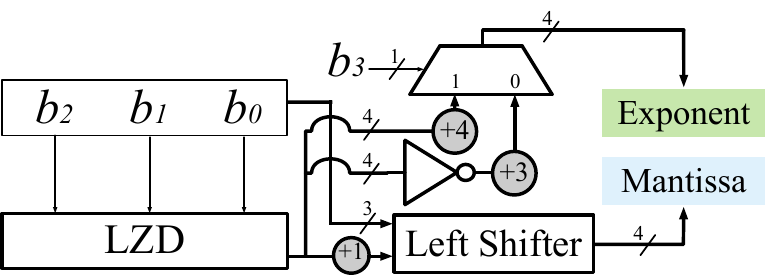}
    \vspace{1mm}
    \caption{The 4-bit unsigned \texttt{float}-based \texttt{flint} decoder.}
    \label{fig:float_flint_decoder}
    \end{center}
\end{figure}

\paragraph{Decoder}
To decode \texttt{int} (\texttt{PoT}) to \texttt{float}, we can set the exponent (mantissa) to zero and copy all bits to mantissa (exponent).
The decoding of \texttt{flint} is more complicated because its decoding is value-dependent.
Thus, we design an efficient \texttt{float}-based \texttt{flint} decoder to address this issue.

The 4-bit \texttt{float}-based unsigned \texttt{flint} design is illustrated in \Fig{fig:float_flint_decoder}, and an arbitrary n-bit \texttt{flint} decoder can be designed in a similar way.
The decoder uses a leading-zero detector (LZD)~\cite{oklobdzija1994algorithmic} and shifters, which are well-known hardware components and both have lightweight implementations.
We use the following equations to extract the exponent and mantissa field from \texttt{flint} type:
\begin{equation}
    \text{Exponent}=\left\{
    \begin{array}{lcl}
        3 - \text{\texttt{LZD}}(b_2b_1b_0), & & b_3 = 0\\
        4 + \text{\texttt{LZD}}(b_2b_1b_0), & & b_3 = 1
    \end{array} \right.,
    \label{eqn:flint_float_exp}
\end{equation}
\begin{equation}
    \text{Mantissa}=b_2b_1b_0 << (\text{\texttt{LZD}}(b_2b_1b_0) + 1)
    \label{eqn:flint_float_int}
\end{equation}
where the \texttt{LZD} is the leading zero number function and $<<$ represents left shift. We can decode \texttt{flint} to the original exponent and mantissa. Finally, as shown in~\Tbl{tbl:ant_value}, the \texttt{float} decoder will continue to transfer them to real values.  For example, a \texttt{flint} number \texttt{1110}$_2$ is \texttt{12}$_{10}$. Its exponent is $4 + \text{\texttt{LZD}}(110) = 4$. Its mantissa is $110 << (0 + 1) = 100_2 = 0.5_{10}$. Therefore, \texttt{1110}$_2$ is $2^{4 - 1} \times 1.5 = 12_{10}$. 

\begin{table}[b]
    \centering
    \ra{1.5}
      \begin{tabular}{c|c|c|c}
        \Xhline{1.2pt}
        Binary   & Exponent & Base Integer & Integer Value \\\Xhline{1.2pt}
        \bluec{0}xxx& 0&0, 1, 2, ..., 7&0, 1, 2, ..., 7  \\ \hline
        \bluec{11}xx& 0&8, 10, 12, 14 &8, 10, 12, 14   \\ \hline
        \bluec{101}x& 2&4, 6&$4 << 2 = 16$, $6 << 2 = 24$\\ \hline
        \bluec{1001}& 4&2 &$2 << 4 = 32$\\ \hline
        \bluec{1000}& 6&1&$1 << 6 = 64$\\\Xhline{1.2pt}
        \end{tabular}
        \caption{\texttt{Int}-based \texttt{flint} 4-bit value table. The blue numbers are the first-one-encoded exponent and ``x'' is 0/1.}
        \label{tbl:flint_int}    
\end{table} 

\subsection{Integer-based PE}
\label{subsec:int_pe}
For DNN inference, it is more common to use \texttt{int}-based PE which is simpler and more area efficient than \texttt{float}-based PE.
Because of the incompatibility between \texttt{int} and \texttt{float}, we remove the latter from the \proj{} primitive data types, which now include \texttt{flint}, \texttt{int}, and \texttt{PoT}.
To support \proj{} on the integer-based PE, we first introduce a unified representation that is based on two \texttt{int} values and its corresponding decoder design. 
We then present the light-weight modification of the original \texttt{int} MAC to support other primitive types in \proj{} such as \texttt{flint} and \texttt{PoT}.

\paragraph{Decoder} 
For a given integer $i$, we decompose it to a base integer $b_i$ and an exponent integer $e$, such that $i = b_i << e$.
\Tbl{tbl:flint_int} shows such a decomposition for 4-bit \texttt{flint} type.
When the most significant bit (MSB) of \texttt{flint} is 0, the base integer value matches the value in \texttt{int} format and the exponent is zero.
When the most significant bit is 1, the base integer value matches the value of remaining bits in \texttt{int} format left-shifted by one, and the exponent can be derived by using the leading-zero detection function/logic.

\begin{figure}[t]
    \begin{center}
    \includegraphics[width=0.9\columnwidth]{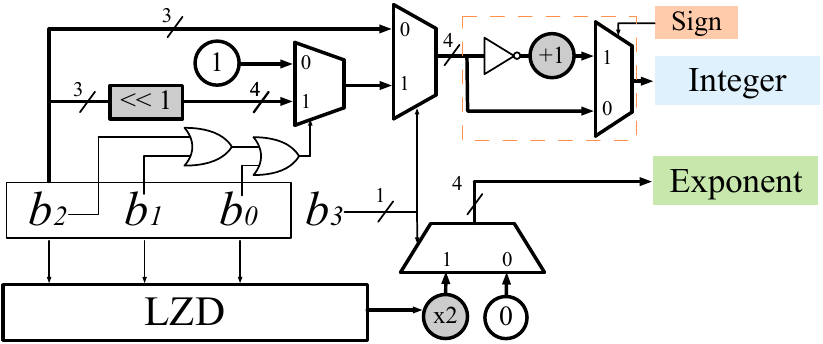}
    \vspace{1mm}
    \caption{\revise{The 4-bit \texttt{int}-based \texttt{flint} decoder.}}
    \label{fig:int_flint_decoder}
    \end{center}
\end{figure}

The following equations describe the base integer and exponent integer for a $flint$ number $x = b_3b_2b_1b_0$. \Fig{fig:int_flint_decoder} illustrates the corresponding decoder design.
\begin{equation}
    \text{Base Integer}=\left\{
    \begin{array}{lcl}
        b_2b_1b_0&, & {b_3 = 0}\\
        b_2b_1b_0 << 1&, & {b_3 = 1}\\
        1&, & {x \ \ = 1000_2}\\
    \end{array} \right. ,
    \label{eqn:unsigned_flint_int}
\end{equation}
\begin{equation}
    \text{Exponent}=\left\{
    \begin{array}{lcl}
        0&, & {b_3 = 0}\\
        2 \times \text{\texttt{LZD}}(b_2b_1b_0)&, & {b_3 = 1}
    \end{array} \right. ,
    \label{eqn:unsigned_flint_exp}
\end{equation}
This representation also works for \texttt{int} and \texttt{PoT}.
The \texttt{int} type has a zero exponent value, while the \texttt{PoT} type has the base integer of one and the exponent value from its binary.

In summary, the \texttt{flint} type expands the value range if \texttt{int} type by using a simple left shifter instead of the complicated hardware logics in \texttt{float} type.
It can be decoded to two integer numbers instead of a fraction number.
Combined with a proper scale factor, the \texttt{flint} can be coupled with \texttt{int} quantization without extra overhead.
The decoder for \texttt{flint} with an arbitrary bit-width can be generated in a similar way. The sign bit can be easily combined with the decoded numbers to fit the \texttt{int} multiplication.

\begin{figure}[t]
    \begin{center}
    \includegraphics[width=0.9\columnwidth]{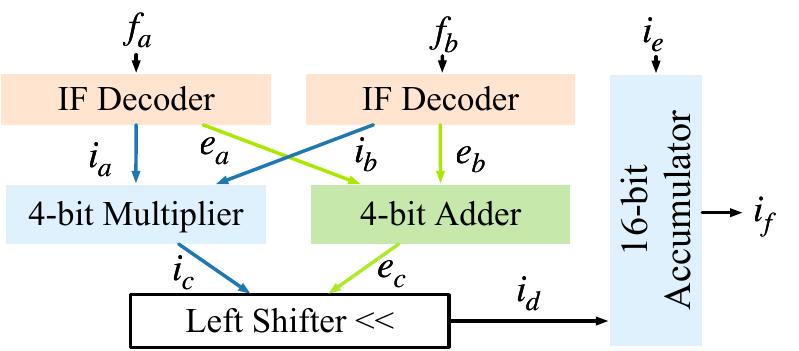}
    \vspace{1mm}
    \caption{The 4-bit \texttt{int}-based \texttt{flint} MAC unit. ``IF Decoder'' is the \texttt{int}-based \texttt{flint} decoder.}
    \label{fig:flint_mac}
    \end{center}
\end{figure}

\paragraph{Multiplier and Accumulator} 
The decoded \texttt{flint} is not directly compatible with the original \texttt{int}-based MAC because of the extra exponent and shift operations. 
Therefore, we need an adder and shifter for \texttt{flint} computation shown in \Fig{fig:flint_mac}.
Assume we have two \texttt{flint} numbers, $f_a$ and $f_b$, with exponent $e_a$ and $e_b$ and base integer $i_a$ and $i_b$, respectively. 
The integer multiplication is the same as original \texttt{int}, including the sign bit, i.e., $i_c = i_a \times i_b$.
The exponent need an add operation $e_c = e_a + e_b$. 
Then, we get the final result $i_d = i_c << e_c$, which can be represented by a 16-bit \texttt{int} number.
As we have explained in \Sec{subsec:ant_quantization}, low-bit \texttt{int} MAC usually adopts a high precision accumulator to achieve the precise accumulation results~\cite{jacob2018quantization, a100}. 
The \texttt{flint} type produces a 16-bit \texttt{int} result and is compatible with the original 16-bit accumulator with $i_f = i_e + i_d$.

\subsection{\revise{Signed Number Support}}
\label{subsce:signed}
The \texttt{flint} decoder function and hardware design are generally extensible for arbitrary bit-width with specific constant settings. 
In particular, we show that the signed number decoder can reuse most of the components in the unsigned number decoder without affecting its critical path.

For example, assume that we have a 4-bit signed \texttt{flint} number. The most significant bit $b_3$ is the sign, and the last three bits are $b_2b_1b_0$. For \texttt{int}-based \texttt{flint}, the following equations are the base integer and exponent decoder for 3-bit \texttt{flint}. Obviously, we can easily reuse the 4-bit unsigned \texttt{flint} decoder function shown in Equation~\eqref{eqn:unsigned_flint_int} and~\eqref{eqn:unsigned_flint_exp}.
\begin{equation}
    \text{Base Integer}=\left\{
    \begin{array}{lcl}
        b_1b_0&, & {b_2 = 0}\\
        b_1b_0 << 1&, & {b_2 = 1}\\
        1&, & {x \ \ = 100_2}\\
    \end{array} \right. ,
    \label{eqn:3bit_signed_flint_int}
\end{equation}
\begin{equation}
    \text{Exponent}=\left\{
    \begin{array}{lcl}
        0&, & {b_2 = 0}\\
        2 \times \text{\texttt{LZD}}(b_1b_0)&, & {b_2 = 1}
    \end{array} \right. ,
    \label{eqn:3bit_signed_flint_exp}
\end{equation}

To maintain the compatibility with the signed integer MAC unit, we need to convert the base integer to its two's complement form as shown in \Fig{fig:int_flint_decoder}.
However, this conversion process does not affect the critical path of the unsigned \texttt{flint} decoder as the critical path still lies in the leading zero detector unit.
For the float-based \texttt{flint}, we can attach the sign bit to the decoded exponent and mantissa based on the original unsigned \texttt{float}-based decoder.

\subsection{Mixed-precision Support}
\label{sec:mixed}

In this work, we propose to couple our \proj{} with the mixed-precision quantization to achieve the same accuracy of the original high-precision DNN models.
According to many prior works~\cite{zhou2016dorefa, micikevicius2018mixed, wang2019haq, cai2020zeroq}, the 8-bit \texttt{int} is sufficient to maintain the original model accuracy.
We explain how our 4-bit \proj{} PE design can naturally support 8-bit \texttt{int} PE.

\begin{figure}[t]
    \begin{center}
    \includegraphics[width=.95\columnwidth]{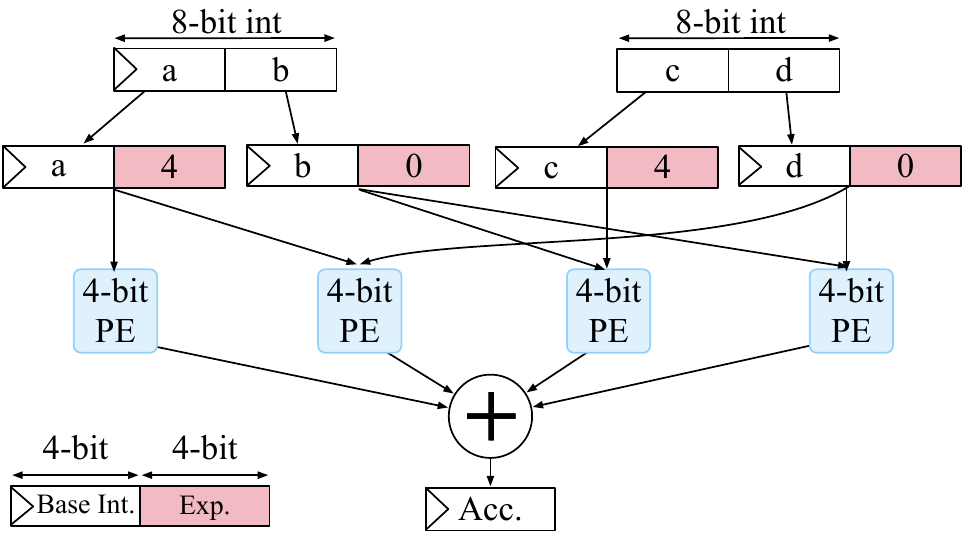}
    \vspace{1mm}
    \caption{The 8-bit \texttt{int} MAC implementation via four 4-bit \proj{} MACs, which reuses most components except the adder.}
    \label{fig:mixed}
    \end{center}
\end{figure}

\Fig{fig:mixed} shows how to use four 4-bit \proj{} PEs to multiply two 8-bit \texttt{int} numbers.
First, we decode the two 8-bit numbers $<a,b>$ and $<c,d>$ to four numbers in our base integer and exponent representation, which are $<a,4>$, $<b,0>$, $<c,4>$, and $<d,0>$. 
Then, we perform four parallel multiplication for those four numbers, as illustrated in \Fig{fig:mixed}, each using a 4-bit \proj{} PE.
Finally, we sum the results of four multiplication using an extra adder tree.
In summary, our \proj{} PE is a good fit for supporting the mixed-precision DNN inference.
In the later evaluation, we show that most tensors (up to 91\%) would use 4-bit \proj{} while only a fraction of tensors would use 8-bit \texttt{int}.

\vspace*{0.2cm}
\section{Architecture Integration}
\label{sec:arch}
In this section, we describe how to integrate the aforementioned \proj{} processing element into existing DNN accelerator architectures such as systolic array and tensor core~\cite{a100}.
We present our optimizations to minimize the overhead of using \proj{}'s TypeFusion PE.
In the end, we describe the convenience of extending the instruction set for our design.

\subsection{ANT and Dataflow Co-design}

\begin{figure}[t]
    \begin{center}
    \includegraphics[width=1\columnwidth]{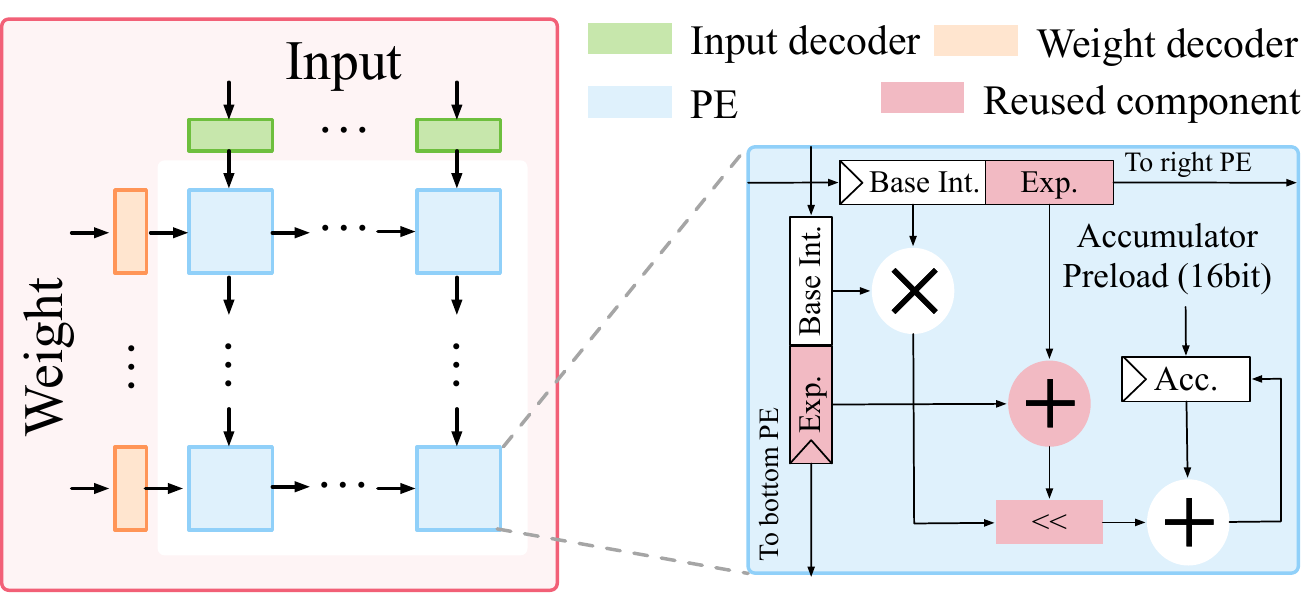}
    \vspace{1mm}
    \caption{Architectural optimizations for integrating \proj{} data type to the output-stationary systolic array.}
    \label{fig:opt_arch}
    \end{center}
\end{figure}

We first describe the architectural optimizations for applying \proj{} to the systolic array, which is also adopted by commercial DNN accelerators like Google's TPU~\cite{jouppi2017datacenter}.
As we have explained in \Sec{subsec:ant_quantization}, our design follows the common practice in which the input and weight \revise{tensor} have low-bit quantization while the output tensor has high-bit quantization.
As such, we find that our design achieves the best benefits on the systolic array with the output-stationary dataflow~\cite{sharma2018bit}.
Our evaluation results show that the weight-stationary systolic has close benefits as well.
\Fig{fig:opt_arch} depicts an output stationary systolic array with \proj{} decoders.

\paragraph{Decoder Placement}
We place \proj{} decoders between the on-chip memory buffer and systolic array. 
This means that quantized tensors are stored with low-bit precision in both off-chip  and on-chip buffers.
Meanwhile, there is no special hardware requirement for off-chip memory accesses because \proj{} numbers are decoded before they enter the systolic array.
This design decision improves both the performance and energy efficiency because BERT-like models are bounded by the off-chip memory bandwidth~\cite{wang2021spatten} while CNN models spend the most energy on on-chip buffer accesses~\cite{chen2016eyeriss}.

As \Fig{fig:opt_arch} shows, only the boundary PEs in the systolic array access the on-chip buffer.
As such, we only place the decoders along the boundary to mitigate the area overhead.
For the output-stationary systolic array, the input and weight elements are sent to the PE (process element) array from the top and left, respectively. 
Assuming the array size of $n\times n$, we only need $2n$ instead of $n^2$ decoders, which amortizes the hardware area overhead of our design. 
The weight-stationary systolic array only needs $n$ decoders for the input tensor as the output tensor is stored with high precision.

\paragraph{PE Connection}
To use \proj{} PEs in the systolic array, we need extra wires connecting neighbouring PEs.
The reason is that after decoding, an $n$-bit \proj{} type has two $n$-bit binary numbers.
For example, a \texttt{float}-based \proj{} type has an $n$-bit exponent number and an $n$-bit mantissa number, while an \texttt{int}-based \proj{} type has an $n$-bit exponent number and an $n$-bit base integer number.
However, our evaluation results show that the extra overhead for those wires is negligible due to the extremely short distance between PEs.

\paragraph{Component Reuse} 
The 4-bit \texttt{int}-based \proj{} MAC unit uses a 4-bit adder and shifter for adding exponent values and shifting the multiplier result, respectively.
Those extra hardware overheads can be mitigated in the mixed-precision design. 
As we have explained in \Sec{sec:mixed}, the 8-bit \texttt{int} MAC requires four 4-bit \texttt{ANT} PE and a 16-bit adder.
In the 8-bit mode, the $n\times n$ systolic array with 4-bit \proj{} PEs would transform to $n/2 \times n/2$ systolic array with 8-bit \texttt{int} PEs.
In this sense, we claim that \proj{} does not introduce new components for the PE of the mixed-precision systolic array, except for decoders outside the systolic array.

\paragraph{Weight Stationary}
Similar to output stationary, we move the input decoder to the top instead of the inner PE, as shown in \Fig{fig:opt_arch}. Because weight elements are preloaded to the PEs, the weight can be decoded before the preloading. Therefore, the weight decoders only need to decode and store the decoded exponent and integer within each PE. Other optimizations for output stationary can also be used similarly as well.

\paragraph{Tensor Core}
Tensor core already supports mixed precision.
For example, the A100 GPU with Ampere architecture~\cite{a100} provides 624 and 1248 TOPS (tera operations per second) for 8-bit \texttt{int} and 4-bit \texttt{int}, respectively.
Meanwhile, the accumulator width for those MAC units is 32-bit \texttt{int}.
As such, the existing tensor core can easily adopt the \proj{} type by augmenting its MAC units and adding decoders for the two multiplication operands.
Moreover, the tensor core-based \proj{} has aligned memory accesses and does not require any modification of GPUs' memory hierarchy.

\subsection{Instruction Set Extension}
\label{subsec:inst}
The \proj{} framework \revise{introduces new data types for the multiply-accumulate instructions.}
For \texttt{int}-based \proj{}, we have two new data types, i.e., \texttt{PoT} and \texttt{flint}. 
They have the fixed bit-width so that the original load/store instructions are still applicable and hence remain unchanged.
\revise{
Thus, there is no modification for the memory sub-system.
}

Obviously, \proj{} does also not break the original programming model for convolutional (CONV) and fully connected (FC) layers. 
The specific type for each CONV and FC layer are determined after the quantization, and we can replace the original \texttt{int}-based version with \texttt{flint} or \texttt{PoT} to generate the corresponding codes.
Thus, our \proj{} framework has a broad applicability owing to its ease of integration.

\vspace*{0.2cm}
\section{{Evaluation}}
\label{sec:evaluation}

We evaluate \proj{} in the aspect of model accuracy, performance, area overhead, and energy efficiency in this section.

\begin{table}[b]
    \ra{1.5}
    \resizebox{0.99\columnwidth}{!}{%
    \begin{tabular}{c|c|c|c|c|c}
    \Xhline{1.2pt}
    Type & \multicolumn{3}{c|}{CNN} & \multicolumn{2}{c}{Transformer} \\
    \Xhline{1.2pt}
    \multirow{2}{*}{Model}
    & VGG16 
    & Res.18 / 50
    & Incep.V3
    & ViT
    & BERT \\
    
    & \cite{he2016deep} 
    & \cite{he2016deep}
    & \cite{szegedy2016rethinking}
    & \cite{dosovitskiy2020image}
    & \cite{devlin2018bert} 
    \\ \hline
    \multirow{2}{*}{Dataset} & \multicolumn{4}{c|}{ImageNet} & GLUE \\
    & \multicolumn{4}{c|}{\cite{deng2009imagenet}} & \cite{wang2018glue} \\ \hline
    Acc. (\%) & 73.48 & 69.59 / 75.97 & 77.34 & 80.99 &
    \begin{tabular}[c]{@{}c@{}}84.42 \\ (MNLI)\end{tabular}    \\ 
    \Xhline{1.2pt}
    \end{tabular}%
    }
    \vspace{1mm}
    \caption{Details of evaluated model and dataset.}
    \label{tab:benchmark}
\end{table}

\subsection{{Methodology}}
\label{subsec:method}
\paragraph{Baselines} 
We implement the \proj{} quantization framework in PyTorch~\cite{paszke2019pytorch}.
We evaluate four baselines compared against  \proj{}, including BitFusion~\cite{sharma2018bit}, OLAccel~\cite{park2018energy}, \revise{BiScaled}~\cite{jain2019biscaled}, AdaFloat~\cite{tambe2020algorithm}, and GOBO~\cite{zadeh2020gobo}. 
BitFusion~\cite{sharma2018bit} uses the mixed-precision of 4-bit and 8-bit \texttt{int} types.
\revise{BiScaled~\cite{jain2019biscaled} quantizes the tensors with two scale factors to address different ranges. We take the accuracy results from BiScaled paper and only synthesize the 6-bit BiScaled BPE.}
AdaFloat~\cite{tambe2020algorithm} requires an 8-bit \texttt{float} to maintain the original model accuracy.
OLAccel and GOBO are both outlier-aware quantization. 
We extend OLAccel~\cite{park2018energy} to the Transformer-based models with weight \& activation quantization.
Note that according to the original paper, the first and last layer require 8-bit instead of 4-bit for normal values.
GOBO~\cite{zadeh2020gobo} only quantizes weights, so we only compare  \proj{} against it in the metrics of area and accuracy.

\paragraph{Benchmark}
We use both CNN and Transformer-based models, including computer vision and natural language processing tasks listed in \Tbl{tab:benchmark}.
We exploit the SOTA checkpoint from PyTorch official repository~\cite{paszke2019pytorch}. We report the top-1 accuracies with FP32 in \Tbl{tab:benchmark}.
The evaluated CNN models with the ImageNet dataset~\cite{deng2009imagenet} include VGG-16~\cite{simonyan2014very}, ResNet-18~\cite{he2016deep}, ResNet-50~\cite{he2016deep}, and Inception-V3~\cite{szegedy2016rethinking}.
For Transformer-based models, we evaluate BERT-Base~\cite{devlin2018bert} with eight datasets of the GLUE dataset suite~\cite{wang2018glue}.
Owing to the space limitation, we only present the results on three datasets (MNLI, CoLA, and SST-2), while the other datasets have similar results.
We also evaluate ViT (vision transformer)~\cite{dosovitskiy2020image}, which is a recent Transformer-based model and has achieved excellent results for vision tasks.

\paragraph{Fine-tuning} 
Our \proj{} along with other baselines except BiScaled~\cite{jain2019biscaled} are compatible with quantization-aware training for better accuracies.
To conduct a fair comparison, we strictly set the same hyper-parameters, including number of fine-tuning epochs and learning rate, for all types. 
All variables use 32-bit floating-point (FP32) arithmetic operations to simulate quantization effects~\cite{jacob2018quantization}. 
We generate and inject trainable weights and activation quantization parameters into the computation graph to fine-tune the quantized weights and activations.
To optimize the clipping ranges (i.e., scale factors in Equation~\eqref{equ:quant}), we also employ the straight-through estimator (STE)~\cite{bengio2013estimating} method in the backward propagation based on the quantization framework PACT~\cite{choi2018pact,li2019additive}. 

\paragraph{Accelerator Implementation}
We implement the \proj{} decoder and PE described in \Sec{sec:typefusion} with the Verilog RTL.
We use Synopsys Design Compiler~\cite{kurup2012logic} to synthesize those components with the 28~$nm$ TSMC process, which reports area and static/dynamic power estimation.
We use CACTI~\cite{muralimanohar2009cacti} to estimate the area, latency, and power of memory structures.
For the end-to-end performance evaluation of \proj{} and other baselines, we develop a cycle-accurate simulator based on the DnnWeaver~\cite{sharma2016high}.
We use DeepScaleTool~\cite{sarangi2021deepscaletool} to scale all designs to the 28~$nm$ process for the iso-area comparison.

\subsection{Quantization Accuracy}
\label{subsec:acc}

Since \proj{} uses multiple primitive data types (\texttt{flint}/\texttt{int}/\texttt{float}/\texttt{PoT}), we first study the contribution of each primitive for improving the quantization accuracy.

\begin{figure}[t]
    \begin{center}
    \includegraphics[width=1\columnwidth]{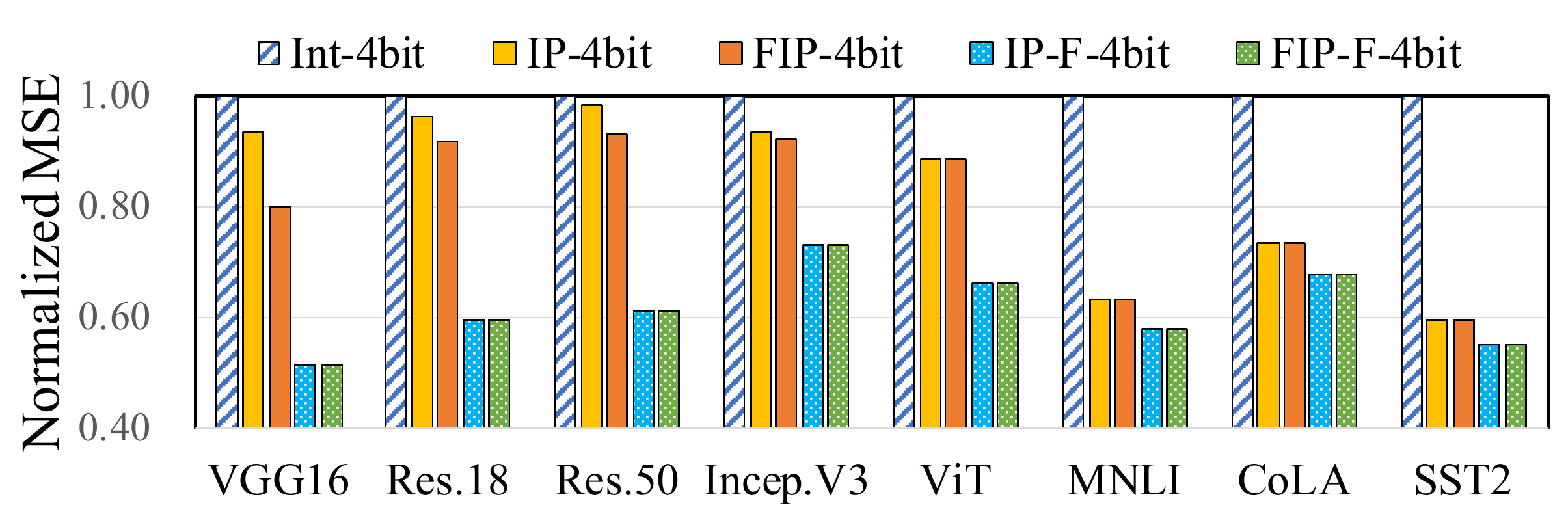}
    \caption{The quantization MSE with the combination of four different primitive types, all of which use 4-bit.}
    \label{fig:mse}
    \end{center}
\end{figure}

\begin{figure}[t]
    \begin{center}
    \includegraphics[width=1\columnwidth]{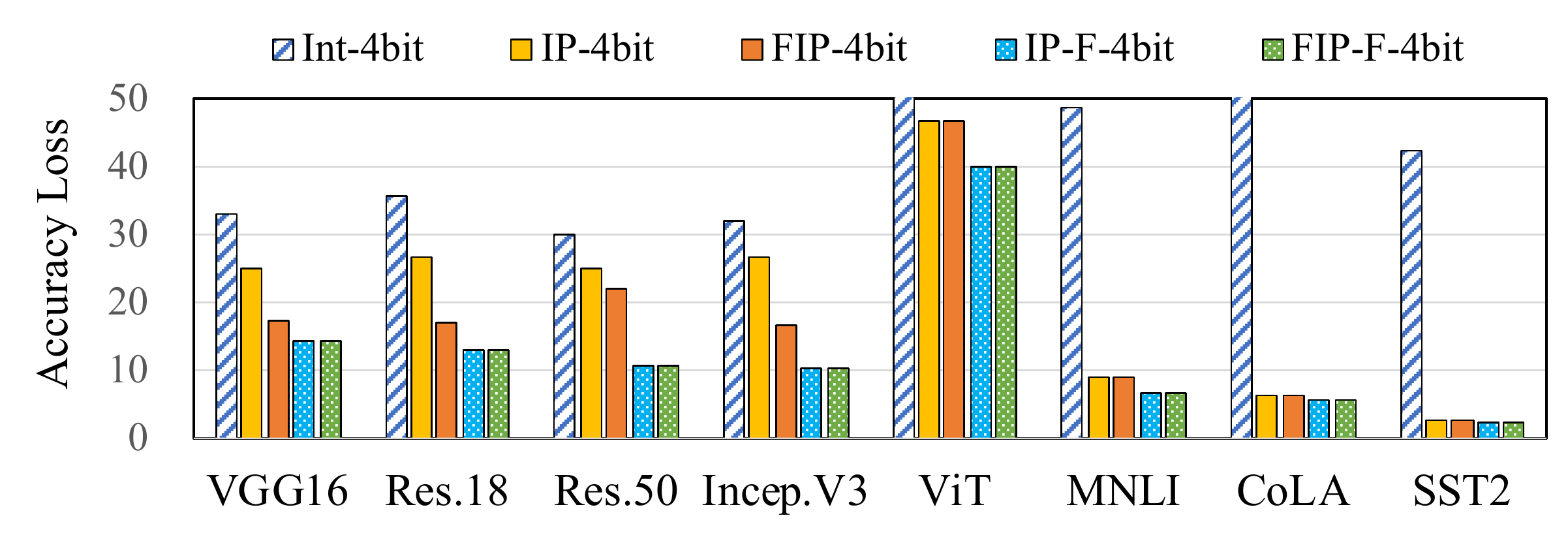}
    \caption{\revise{The accuracy loss without fine-tuning.}}
    \label{fig:ptq-accuracy}
    \end{center}
    \vspace*{-4mm}
\end{figure}

\begin{figure}[t]
    \begin{center}
    \includegraphics[width=1\columnwidth]{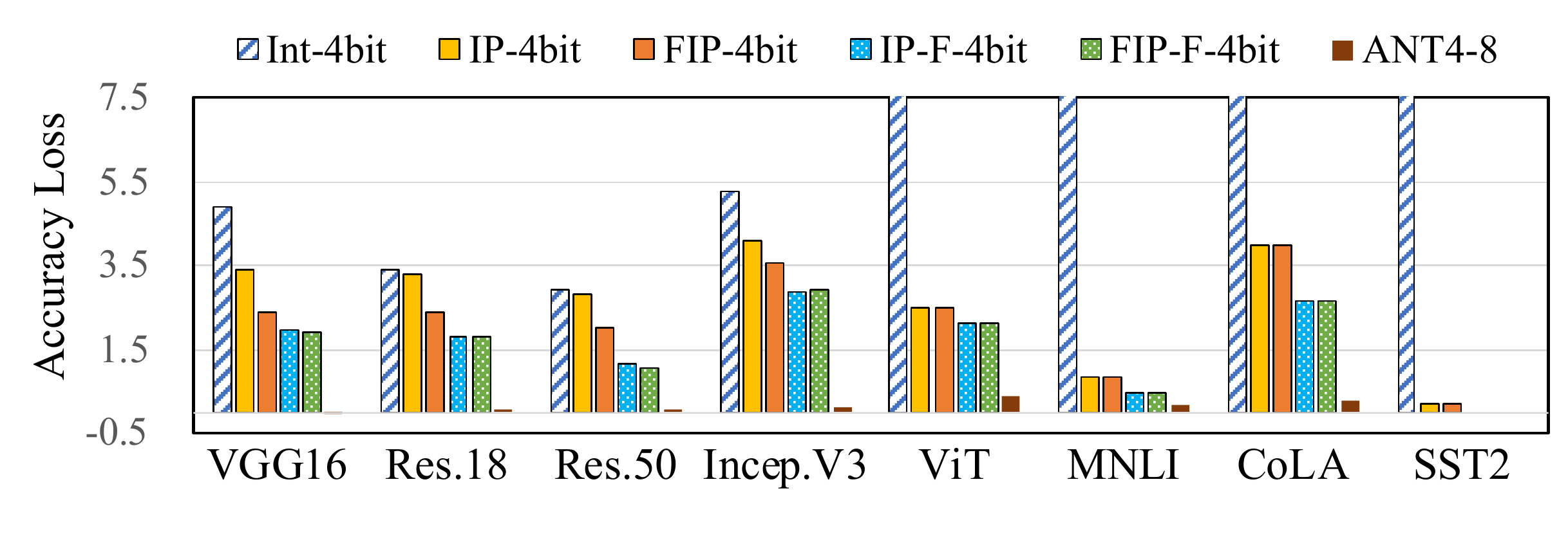}
    \caption{The accuracy loss \revise{with fine-tuning}.}
    \label{fig:accuracy}
    \end{center}
    \vspace*{-4mm}
\end{figure}

\paragraph{Primitive Combination}
We study six combinations of the four primitive data types.
\texttt{Int} is the combination with only a single data type.
Two combinations \texttt{int}-\texttt{PoT} (IP) and \texttt{float}-\texttt{int}-\texttt{PoT} (FIP) excludes \texttt{flint} and hence only exploit the \emph{inter-tensor} adaptivity.
Correspondingly, we evaluate these two combinations that add \texttt{flint} and exploit the \emph{intra- and inter-tensor} adaptivity.
They are \texttt{int}-\texttt{PoT}-\texttt{flint} (IP-F) and \texttt{float}-\texttt{int}-\texttt{PoT}-\texttt{flint} (FIP-F).  \texttt{ANT4-8} uses the mixed-precision of 4-bit \texttt{int}-based \texttt{ANT} (i.e., IP-F) and 8-bit \texttt{int} for accuracy comparison.
All types use 4-bit quantization except \texttt{ANT4-8}.
For the quantization metrics, we use the MSE and model accuracy loss against the original FP32 model.
\Fig{fig:mse} plots the quantization MSE of these combinations on eight DNN models. 
\Fig{fig:ptq-accuracy} and \Fig{fig:accuracy} demonstrate their accuracy loss compared to original high-precision models before and after fine-tuning, respectively.

\paragraph{Quantization MSE}
For quantizing each tensor in DNN models, we employ the \proj{} algorithm described in in \Sec{subsec:ant_quantization} to choose the primitive data type with minimum MSE. 
From the results in \Fig{fig:mse}, we find that adding more primitive data types generally lets us decrease the accuracy loss owing to quantization errors.
In specific, adding the \texttt{PoT} type is critical for Transformer-based models on NLP datasets (MNLI, CoLA, and SST2), since they have large activation values.
The benefit of the \texttt{PoT} type is smaller for the vision tasks including ViT.
Adding the \texttt{flint} type is important for both vision and NLP tasks.
Finally, we observe that adding the \texttt{float} has the least impact on the quantization errors, whose role is replaced by other primitive types.

\begin{table}[b]
    \centering
    \ra{1.5}
    \resizebox{0.75\columnwidth}{!}{%
      \begin{tabular}{c|c|c|c}
        \Xhline{1.2pt}
        Model & \proj{} &  BiScaled & Source \\ \Xhline{1.2pt}
        AlexNet~\cite{krizhevsky2012imagenet}  & \textbf{55.85\%}  &54.90\%   &56.56\% \\ \hline
        VGG16  & \textbf{72.80\%}  & 66.56\%   &73.48\% \\ \hline
        ResNet50 & \textbf{75.08\%}& 70.46\% &75.97\%\\\hline
        ResNet152  & \textbf{77.30\%} &   73.41\%   &78.25\%\\
        \Xhline{1.2pt}
        \end{tabular}
    }
        \vspace{1mm}
        \caption{\revise{Accuracy comparison between \proj{} and BiScaled without fine-tuning under 6-bit quantization.}}
        \label{tbl:biscaled}    
\end{table}

\paragraph{Accuracy}
Comparing \Fig{fig:mse},  \revise{\Fig{fig:ptq-accuracy}, and \Fig{fig:accuracy}}, we find that the model accuracy loss correlates well with the quantization MSE, and the fine-tuning plays an essential role in recovering the accuracy to the original values before quantization.
The 4-bit IP-F and FIP-F provide more numerical types for selection, and both achieve the minimum accuracy loss.
The former only requires the \texttt{int}-based PE while the latter requires the \texttt{float}-based PE.
We show later that the \texttt{float}-based PE for \proj{} consumes almost $3\times$ area of \texttt{int}-based PE.
As such, we choose the IP-F configuration (i.e., \texttt{int}-\texttt{PoT}-\texttt{flint}) as the final \proj{} for the rest of evaluation.
Note that the 4-bit \proj{} type is still not able to maintain the original model accuracy,  which justifies the choice of mixed-precision in our work.
The mixed-precision \texttt{ANT4-8} type can achieve original model accuracy in \revise{CNN} models and less than $1\%$ accuracy loss for ViT and BERT. 
We also observe that our proposed \texttt{flint} data type is important for the accuracies of both vision and NLP tasks.
Meanwhile, the \texttt{PoT} type is more important for Transformer-based models on NLP tasks than vision tasks.

\paragraph{Comparison against BiScaled}
We first compare the accuracy of IP-F configuration (i.e., \texttt{int}-\texttt{PoT}-\texttt{flint}) of \proj{} against the BiScaled~\cite{sharma2018bit} without fine-tuning.
\Tbl{tbl:biscaled} shows the 6-bit quantization without fine-tuning results for \proj{} and BiScaled. We find that \proj{} offers much better accuracy than BiScaled because \proj{} can exploit inter-tensor adaptivity and intra-tensor adaptivity with more exponent domains. 

\begin{table}[b]
    \centering
    \ra{1.5}
    \resizebox{0.85\columnwidth}{!}{%
      \begin{tabular}{c|c|c|c}
        \Xhline{1.2pt}
        Bit Width &\proj{} & GOBO & Source \\ 
        \Xhline{1.2pt}
        3-bit& \textbf{83.86\%}  & 83.76\% (3.04 bit) &\multirow{2}{*}{84.42\%} \\\cline{1-3}
        4-bit& 84.39\% &\textbf{84.45\%} (4.04 bit) &  \\ 
        \Xhline{1.2pt}
        \end{tabular}
    }
        \vspace{1mm}
        \caption{Accuracy  comparison between weight-only quantization using \proj{} and GOBO for BERT on MNLI dataset.}
        \label{tbl:gobo}    
\end{table}

\begin{figure*}[t]
    \begin{center}
    \includegraphics[width=1\textwidth]{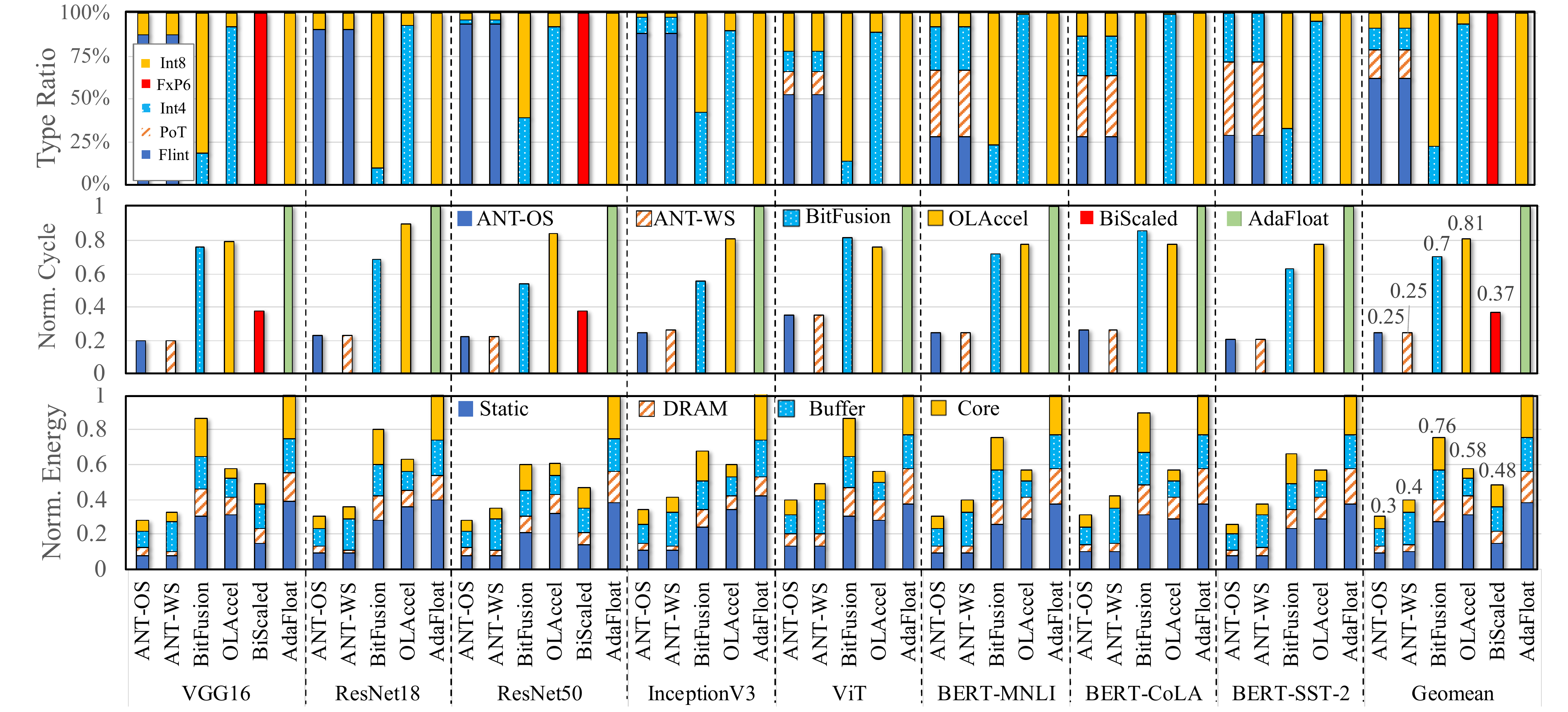}
    \caption{\revise{Comparison of the tensor type ratios (top), normalized latency (middle), and energy in different designs (bottom).}}
    \label{fig:all_data}
    \end{center}
    \vspace{-4mm}
\end{figure*}

\paragraph{Comparison against GOBO}
We compare the accuracy of \proj{} against the prior outlier-aware quantization work GOBO~\cite{zadeh2020gobo}.
Unlike \proj{} that performs both weight and activation quantization, GOBO only performs weight quantization.
For a fair comparison, \Tbl{tbl:gobo} shows that the weight-only quantization using \proj{} achieves a similar accuracy, while \proj{}'s fixed-length feature is more hardware-friendly than the GOBO's variable-length encoding scheme.

\subsection{{Area}}
According to our evaluation, the \texttt{float}-based PE has about $3\times$ area of \texttt{int}-based PE. 
Given their similar accuracies, we choose to use the \texttt{int}-based decoder and PE for \proj{} accelerator. 
We compare the accelerator area breakdown in \Tbl{tab:area}. 
Overall, the \texttt{int}-decoder overhead is about $0.2\%$ for the systolic array. 
In the rest of evaluation, we scale other accelerators to 28~$nm$ and perform an iso-area comparison. All accelerators have the same on-chip buffer configuration.

\begin{table}[b]
    \ra{1.5}
    \resizebox{\columnwidth}{!}{%
    \begin{tabular}{c|l|c|c|c}
    \Xhline{1.2pt}
    \multirow{2}{*}{Architecture} & \multicolumn{3}{c|}{Core} & \multirow{2}{*}{Buffer}  \\ \cline{2-4} 

    & Component & Number & Area ($mm^2$) & \\ \Xhline{1.2pt}
    
    \multirow{2}{*}{\proj{}}& Decoder (4.9$\mu m^2$) & 128 &\multirow{2}{*}{0.327} & \\ \cline{2-3}
    & 4-bit PE (79.57$\mu m^2$) & 4096 & &\\ \cline{1-4}

    BitFusion & 4-bit PE & 4096 & 0.326 &512 KB     \\ \cline{1-4}
    OLAccel8 & 4-bit \& 8-bit PE & 1152 & 0.320 & 4.2 $mm^2$   \\ \cline{1-4}
    BiScaled & 6-bit BPE  & 2560 & 0.328 &   \\ \cline{1-4}
    AdaFloat & 8-bit PE & 896 & 0.327 & \\ \cline{1-4}

    \Xhline{1.2pt}
    \end{tabular}%
    }
    \vspace*{1mm}
    \caption{The configuration and area breakdown of \proj{} and other baselines under 28~$nm$ process.}
    \vspace*{-4mm}
    \label{tab:area}
\end{table}

\subsection{Performance and Energy}
\label{subsec:perf}

We implement \proj{} with output-stationary (\texttt{ANT-OS}) and weight-stationary (\texttt{ANT-WS}). 
We adjust the mixed-precision ratio to make all models close to their original accuracy (CNN with $< 0.1\%$ loss and Transformer with $< 1\%$ loss) for the iso-accuracy and iso-area comparison \revise{except BiScaled}. 
\revise{
We only compare BiScaled on VGG16 and ResNet50, which have unignorable ($>5\%$) accuracy loss, as shown in \Tbl{tbl:biscaled}.}
Since AdaFloat~\cite{tambe2020algorithm} does not support mixed-precision, we only conduct 8-bit quantization based on AdaFloat.
\Fig{fig:all_data} compares \proj{} design against various baselines with the metrics including the ratio of tensor types, normalized latency, and energy. 
The batch size is 64 for all experiments.

\paragraph{Tensor Type Ratio} 
The top plot of \Fig{fig:all_data} compares the ratio of 4-bit (\texttt{flint}, \texttt{PoT}, and \texttt{int}) and 8-bit (\texttt{int}) tensors in different designs.
\texttt{ANT-OS} and \texttt{ANT-WS} use the same quantization algorithm but different microarchitectures, so that they have the same ratio of various data types.
By inspecting the tensor ratio, we find that CNN models and vision transformer model ViT choose to use a significant portion of 4-bit \texttt{flint} type, while NLP Transformer models use a roughly same portion of 4-bit \texttt{flint} and \texttt{PoT} type.

Compared to the prior mixed-precision work BitFusion~\cite{sharma2018bit}, \proj{} has a much greater ratio of 4-bit tensors because its inter-tensor and intra-tensor adaptivity make the 4-bit \proj{} achieve much lower quantization errors. 
Especially for BERT on SST-2, \proj{} can get the original accuracy with 100\% 4-bit quantization. 
OLAccel~\cite{park2018energy} is not tensor-wise quantization as it uses variable-length encoding for different values within a tensor.
We show its element-wise ratio of 4-bit and 8-bit values in the plot.
Owing to its fine-grained element-wise quantization, it has a slightly higher proportion of 4-bit values than \proj{}, but also incurs a much greater hardware overhead with low end-to-end latency.

\paragraph{Performance} 
The middle plot of \Fig{fig:all_data} compares the normalized execution time of different designs, which shows that \proj{} achieves the best latency performance.
We also find that \texttt{ANT-OS} and \texttt{ANT-WS} have very similar performances because their architectural differences can be mitigated through tiling optimizations~\cite{sharma2018bit}. 
BitFusion has more 8-bit tensors, which lead to its worse performance.
Even though OLAccel has a higher proportion of 4-bit tensors, it needs the additional outlier controller with significant overhead to orchestrate the computation among normal values and outliers. 
In the end, \proj{} achieves averaged 2.8$\times$, 3.24$\times$, \revise{1.48$\times$}, and $4\times$ speedup over BitFusion~\cite{sharma2018bit}, OLAccel~\cite{park2018energy}, BiScaled~\cite{sharma2018bit}, and AdaFloat~\cite{tambe2020algorithm}, respectively.

\paragraph{Energy}  
The bottom plot of \Fig{fig:all_data} compares the normalized energy consumption of different designs, which includes the static energy and dynamic energy (DRAM, on-chip buffer, and core). 
\texttt{ANT-OS} and \texttt{ANT-WS} have the lowest and second-lowest energy, respectively. 
Even though \texttt{ANT-WS} has a similar performance to \texttt{ANT-OS}, \texttt{ANT-WS} needs more buffer accesses for the high-precision output activation. 
Thus, \texttt{ANT-WS} spends more energy on accessing on-chip buffers.
OLAccel consumes less energy than BitFusion because it has more 4-bit values, which reduces the energy of DRAM and on-chip buffer.
In the end, \texttt{ANT-OS} achieves averaged 2.53$\times$, 1.93$\times$, \revise{1.6$\times$}, and $3.33\times$ energy reduction over BitFusion, OLAccel, \revise{BiScaled}, and AdaFloat, respectively.

\subsection{{ANT Type Selection Analysis}}
\label{subsec:type_dis}

In this subsection, we study the effectiveness of the data type selection algorithm in \proj{}, as previously described in \Sec{subsec:ant_quantization}.
Specifically, we present the MSE for all weight and activation tensors in DNN models with diverse distributions.

\begin{figure}[t]
    \begin{center}
    \includegraphics[width=1\columnwidth]{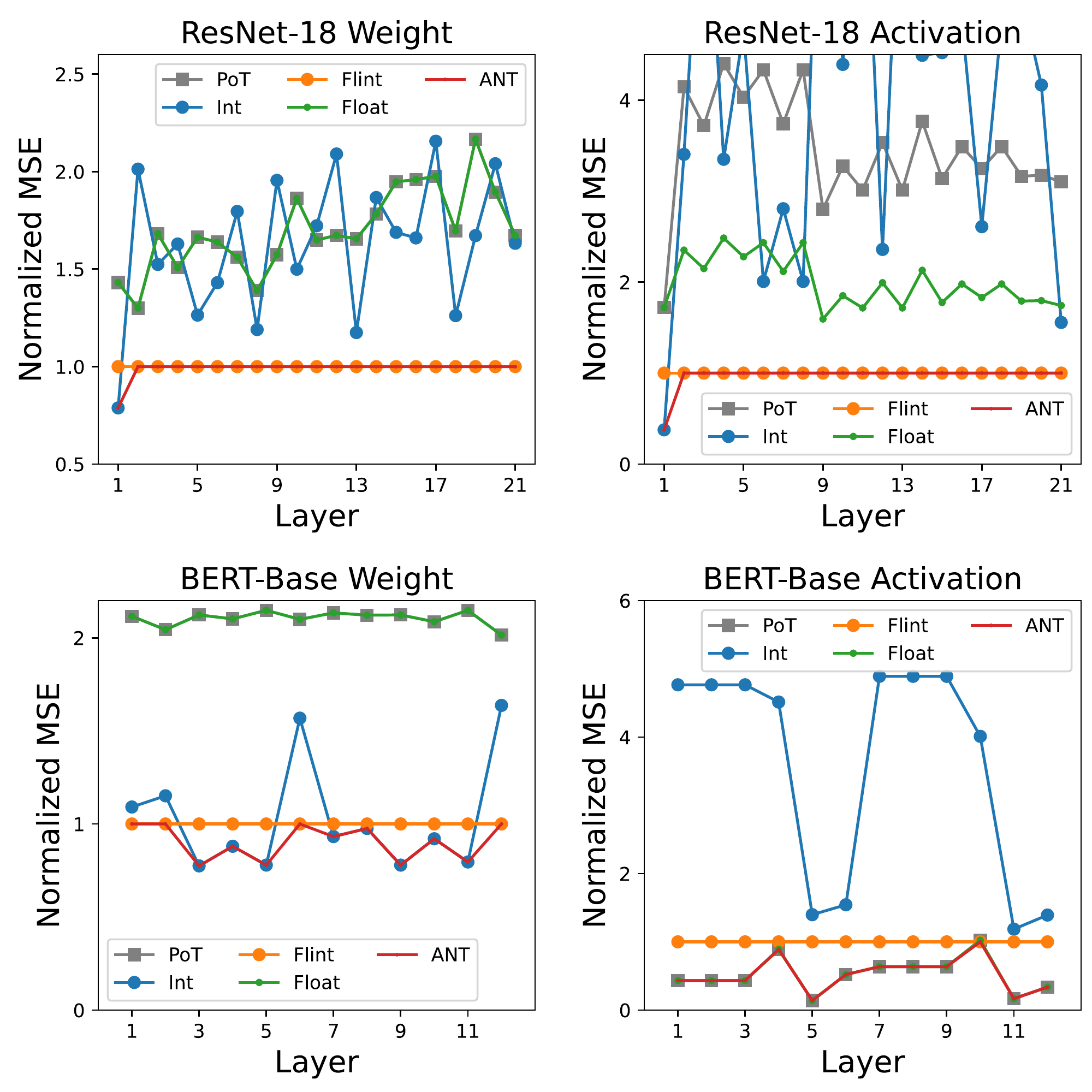}
    \caption{Numerical type (4-bit) mean square error (MSE) results that are normalized to \texttt{flint}.}
    \label{fig:type_distri}
    \end{center}
\end{figure}

We collect weight and activation tensors from ResNet-18 (CNN model) on ImageNet and BERT-Base (Transformer model) on MNLI dataset.
\Fig{fig:type_distri} shows the MSE values of different 4-bit data types that are all normalized to \texttt{flint}.
We adopt the unsigned numerical type for ResNet-18 activation tensors because of the ReLU function, which is a common practice for CNN quantizations~\cite{park2018energy,sharma2018bit,li2019additive,tambe2020algorithm}. 
We use signed types for ResNet-18 weight tensors and all tensors of BERT. 
\Fig{fig:type_distri} shows that \proj \emph{always} chooses the most appropriate data type, i.e., the type with the minimum MSE.
Note that signed 4-bit \texttt{float} and \texttt{PoT} are identical so that they overlap in ResNet-18 weight MSE and BERT-Base weight and activation MSE.

We also justify the choice of data types by inspecting the distribution of different tensors.
Recall that in \Sec{subsec:quant_metrics}, it is natural to choose a numeric type whose quantization resolution distribution is similar to the tensor distribution to reduce the quantization error.
For CNN models, \texttt{int} has pretty low MSEs with the first convolutional layer. 
Our manual inspection shows that the first layer is more like a uniform distribution than Gaussian. 
This is especially so for the activation tensor, which is the original image and not the featured map. 
After the first layer, tensors in CNN models are closer to Gaussian distribution so that \texttt{flint} almost dominates these layers with very low MSE values.

For BERT, we only collect the former two Transformer blocks as the representative, which have a similar trend to the rest Transformer blocks. 
BERT model has relatively more complex tensor distributions. 
The weight tensors show both uniform-like and Gaussian-like distributions so both \texttt{int} and \texttt{flint} are chosen.
On the other hand, activation tensors have significant outliers so that they prefer \texttt{PoT} or \texttt{float}.

\vspace*{0.2cm}
\section{Related Work}

This section presents related work on DNN acceleration, sparse accelerators, and low-bit quantization accelerators.


\paragraph{DNN Acceleration}
To accelerate the DNN models efficiently, researchers proposed both various hardware and software solutions. For hardware acceleration, the proposed architectures are tailored to fit the computation characteristics of DNN models which leverage the regular access pattern, high data reuse and tremendous parallelism to save the area and latency from control logic~\cite{chen2014diannao, chen2014dadiannao, peemen2013memory, zhang2015optimizing, gokhale2014240, gupta2015deep, guo2020balancing, qin2020sigma, gan2020ptolemy, gan2020low, leng2020asymmetric, zhou2021characterizing}.
In these hardware accelerators, weight or weight data flow through multiple stages to maximize reuse, with example like systolic array~\cite{du2015shidiannao, jouppi2017datacenter} and other spatial architectures~\cite{chen2016eyeriss, zhao2019cambricon, nori2021reduct, yuan2021forms}.
Modern GPUs have already deployed SIMD-friendly matrix-matrix multiplication (GEMM) accelerator like tensor core~\cite{a100}.

For software acceleration, the efforts are mainly put into the compilation and scheduling optimizations. To fully utilize the hardware resources, various automated compilers or graph optimizers are proposed to find optimal implementations on different hardware~\cite{TVM, FlexTensor, Ansor, Roller, TASO}. Researchers proposed various scheduling techniques~\cite{Baymax, Prophet, Ebird, LazyBatch, DVABatch, BubbleUp, BubbleFlux, Heracles, VELTAIR} to manage resource usage, task queuing, runtime batching, and so on. 

\paragraph{Sparse DNN Accelerators}
Given the increasing computation demand of DNN models, it is of paramount importance to leverage the algorithm and hardware co-design.
Researchers have proposed pruning and quantization methods to exploit the redundancy property of DNNs for such a purpose.
Pruning means removing part of the weight, input, or even output of DNN layers, which leads to a sparse model with a portion of model size.
However, sparse models contain irregular memory accesses, which could negate the benefits of sparsity.
To overcome this challenge, it is important to design sparsity-optimized algorithms and hardware architectures~\cite{han2015deep, albericio2016cnvlutin, zhang2016cambricon, zhou2018cambricon,  zhu2019sparse,Qiu_2019_CVPR, guan2020far, qin2020sigma, guo2020accelerating,  wang2021dual, guan2022block, guan2022transkimmer}. 

\paragraph{Quantization Accelerators}
The DNN model quantization exploits the insight that DNN inference does not need  high-precision representations like FP32, and is orthogonal to model pruning.
It uses a narrow bit width to reduce memory and computation requirement.
The fixed-length value encoding is convenient for architectural integration because it only requires processing element design, such as FP16, INT8, or even INT4~\cite{sharma2018bit, park2018energy, zadeh2020gobo, song2020drq, jouppi2017datacenter,a100}, and BF16~\cite{jouppi2017datacenter}, TF32~\cite{a100}, and Posit~\cite{gustafson2017beating}. 
Posit is a general data type and a potential replacement for IEEE 754~\cite{kahan1996ieee}. It uses variable length encoding for the regime bits to extend the exponent range. 
Our proposed \texttt{flint} is different from Posit in the aspect that \texttt{flint} has no regime bit and an efficient encoding/decoding process based on \texttt{float} or \texttt{int} type. 

BitFusion~\cite{sharma2018bit} and DRQ~\cite{song2020drq} can support different bit-width via a spatial and temporal combination of low-bit PEs, respectively.
There are also outlier-aware quantization accelerator designs, such as OLAccel~\cite{park2018energy}, DRQ~\cite{song2020drq}, and GOBO~\cite{zadeh2020gobo}, which are more aggressive and require heavy architectural modifications.
Moreover, these outlier-aware quantization accelerators have unaligned computation and memory accesses, resulting in their limited benefits.
In contrast, our work provides the adaptive numerical data type, which provides low-bit fixed-length value presentations and hence is also easy for architectural integration.

\paragraph{Quantization Methods}
In our work, we use two popular quantization methods, i.e., quantization-aware training (QAT)~\cite{gupta2015deep, jacob2018quantization, wang2019learning, zhuang2021effective} and post-training quantization (PTQ)~\cite{gupta2015deep, jacob2018quantization, wang2019learning, zhuang2021effective, guo2022squant}.
The QAT requires fine-tuning to restore the model accuracy, while the latter leverages heuristics such as constraint optimization to avoid fine-tuning.

\vspace*{0.2cm}
\section{Conclusion}

In this work, we present a novel, composite data type called \proj{} to achieve low-bit quantization for accelerating DNN models. The key insight is adapting the data type to value importance within a tensor and different tensors' value distributions.
For the intra-tensor adaptivity, we propose \texttt{flint}, a new data type that combines the advantages of \texttt{int} (maintaining a high precision for important value ranges) and \texttt{float} (maintaining a large value range). 
For the inter-tensor adaptivity, we propose the composite \proj{} type, which selects a data type (e.g., \texttt{int}/\texttt{flint}/\texttt{PoT}) for each tensor according to its distribution. 
We design a unified processing element architecture for \proj{} and show its ease of integration to existing DNN accelerators. Our design demonstrates $2.8\times$ latency reduction and $2.5\times$ energy improvement over the state-of-the-art quantization accelerators.

\vspace*{0.2cm}
\section*{Acknowledgment}
This work was supported by the National Key R\&D Program of China under Grant 2021ZD0110104, the National Natural Science Foundation of China (NSFC) grant (U21B2017, 62072297, and 61832006).
The authors would like to thank the anonymous reviewers for their constructive feedback for improving the work. 
We also thank Tailong Wangliu, Weiming Hu, and Yuxian Qiu for their technical supports and beneficial discussions.

\appendix
\section{Artifact Appendix}
\subsection{Abstract}

Our experiments have two major parts: the evaluation of DNN model accuracy and the performance of the ANT simulator. 

We evaluate the results with models in image classification and NLP. The image classification tasks include five models, i.e., VGG16, ResNet18, ResNet50, Inception-V3, and ViT. We adopt the BERT model for the NLP task with three datasets, MNLI, CoLA, and SST-2. We provide the fine-tuning source code for all models to measure the accuracy. However, that may need dozens of hours to complete the fine-tuning process. Therefore, we provide the checkpoints for the fast evaluation of image classification models, which can finish in one hour. 
For measuring the performance, we evaluate all models with six simulator configurations. In all experiments, we run those models according to the experiment setup on a Ubuntu server that equips an NVIDIA A100 GPU and multiple servers with four NVIDIA A10 GPUs for distributed fine-tuning.

\subsection{Artifact check-list (meta-information)}

{\small
\begin{itemize}
  \item {\bf Compilation: } NVCC 11.3, GCC 7.5.0.
  \item {\bf Model: }VGG-16, ResNet-18, ResNet-50, Inception-V3, ViT, and BERT-Base.
  \item {\bf Data set: }ImageNet, and GLUE dataset.
  \item {\bf Run-time environment: }Ubuntu 18.04.5 LTS, CUDA 11.3, and PyTorch 1.11.
  \item {\bf Hardware: }A server with an x86 processor, an NVIDIA A100 GPU, and a server with four NVIDIA A10 GPUs.
  \item {\bf Output: }Model accuracy, simulator energy, and performance.
  \item {\bf How much disk space required (approximately)?: } 20GB.
  \item {\bf How much time is needed to prepare workflow (approximately)?: } It takes about 30 minutes to prepare the environment.
  \item {\bf How much time is needed to complete experiments (approximately)?: } It takes approximately 50 hours to execute all experiments using the server equipped with GPUs. The fast evaluation can take only one hour with the checkpoints.
  \item {\bf Publicly available: } Our framework is publicly available on GitHub \url{https://github.com/clevercool/ANT_Micro22}.
  \item {\bf Code licenses:} Apache-2.0 license.
  \item {\bf Data licenses:} The datasets are publicly available through their original licensing terms.
  \item {\bf Archived: } \url{https://doi.org/10.5281/zenodo.7002114}.
\end{itemize}
}

\subsection{Description}

\subsubsection{How to access}

We archive the source code at \url{https://doi.org/10.5281/zenodo.7002114}. We recommend you access our GitHub repository: \url{https://github.com/clevercool/ANT_Micro22} for the latest version.

\subsubsection{Hardware dependencies}
We fine-tune the DNN models with two types of server configuration:
A server is equipped with a single NVIDIA A100 (40GB) GPU, and a server is equipped with four NVIDIA A10 (24GB) GPUs for distributed fine-tuning.
\subsubsection{Software dependencies}
The experiments rely on the following software components.
\begin{itemize}
    \item Ubuntu 18.04.5 LTS
    \item Python 3.8
    \item PyTorch 1.11
    \item Andconda 4.10.1
    \item GCC 7.5.0
    \item CUDA 11.3
    \item Cacti 7.0
\end{itemize}
\subsubsection{Data sets and models}
The evaluated image classification models with the ImageNet dataset~\cite{deng2009imagenet} include VGG-16~\cite{simonyan2014very}, ResNet-18~\cite{he2016deep}, ResNet-50~\cite{he2016deep}, Inception-V3~\cite{szegedy2016rethinking}, and ViT (vision transformer)~\cite{dosovitskiy2020image}.
For NLP models, we evaluate BERT-Base~\cite{devlin2018bert} with the GLUE dataset suite~\cite{wang2018glue}.
Owing to the space limitation, we only present the results on three datasets (MNLI, CoLA, and SST-2).

\subsection{Installation}

We have  well-documented README files to detail the installation instruction for each experiment at 
\url{https://github.com/clevercool/ANT_Micro22}.

\subsection{Evaluation and expected results}

Our experiments have two major parts: the evaluation of DNN model accuracy and the performance of the ANT simulator.

\begin{itemize}
    \item The directory \benchmark{ant\_quantization} contains the ANT framework based on PyTorch for the DNN model accuracy evaluation.
    \item The directory \benchmark{ant\_simulator} contains the performance and energy evaluation of the ANT simulator.
\end{itemize}

To evaluate the experiments, you can utilize the scripts in each directory according to the README files.
We also release all expected results in the README files for Figure~\ref{fig:accuracy}, Figure~\ref{fig:all_data}, and Table~\ref{tbl:biscaled}.

\subsection{Methodology}

Submission, reviewing and badging methodology:

\begin{itemize}
  \item \url{https://www.acm.org/publications/policies/artifact-review-badging}
  \item \url{http://cTuning.org/ae/submission-20201122.html}
  \item \url{http://cTuning.org/ae/reviewing-20201122.html}
\end{itemize}

\bibliographystyle{IEEEtranS}
\bibliography{paper}

\end{document}